%% file: arxiv_main.tex
\documentclass{article}

\PassOptionsToPackage{numbers, compress}{natbib}

    \usepackage[final]{neurips_2025}

\usepackage[utf8]{inputenc} 
\usepackage[T1]{fontenc}    
\usepackage{hyperref}       
\usepackage{url}            
\usepackage{booktabs}       
\usepackage{amsfonts}       
\usepackage{amsmath}        
\usepackage{nicefrac}       
\usepackage{microtype}      
\usepackage{xcolor}         
\usepackage{graphicx}
\usepackage{threeparttable}
\usepackage{multirow}
\usepackage{tabularx}
\usepackage{pifont}
\usepackage{wrapfig}

\usepackage{graphicx}
\usepackage{float}
\usepackage{algorithm2e}
\RestyleAlgo{ruled} 
\usepackage{caption}
\usepackage{booktabs}
\usepackage{adjustbox}
\usepackage{makecell} 
\usepackage{enumitem}
\usepackage{graphicx}
\usepackage{float}
\usepackage{algorithm2e}
\RestyleAlgo{ruled} 
\usepackage{caption}
\usepackage{booktabs}
\usepackage{adjustbox}
\usepackage{makecell} 
\usepackage{listings}
\usepackage{tabularx}
\usepackage{adjustbox} 
\usepackage[most]{tcolorbox}
\newcommand{\ourmethod}{Deep Video Discovery }

\def\AdaReTaKe{AdaR\scalebox{0.8}{E}T\scalebox{0.8}{A}K\scalebox{0.8}{E}}

\title{\ourmethod: Agentic Search with Tool Use for Long-form Video Understanding}

\author{
Xiaoyi Zhang\thanks{Equal contribution.
}~~$^{1}$ \quad 
Zhaoyang Jia$^{*2}$\thanks{This work was done during the internship at Microsoft Research Asia as an open-source project.} \quad 
Zongyu Guo$^{1}$ \\
\textbf{Jiahao Li$^{1}$} \quad \textbf{Bin Li$^{1}$} \quad \textbf{Houqiang Li$^{2}$} \quad \textbf{Yan Lu$^{1}$} \\[0.5em]
$^{1}$Microsoft Research Asia \\
$^{2}$University of Science and Technology of China \\
\texttt{\footnotesize   \{xiaoyizhang, zongyuguo, jiahaoli, binli, yanlu\}@microsoft.com}
\\
\texttt{\footnotesize   \{jzy\_ustc, lihq\}@ustc.edu.cn}
\vspace{-0.5cm}
}

\begin{document}

\maketitle

\begin{abstract}
Long-form video understanding presents significant challenges due to extensive temporal-spatial complexity and the difficulty of question answering under such extended contexts. 
While Large Language Models (LLMs) have demonstrated considerable advancements in video analysis capabilities and long context handling, they continue to exhibit limitations when processing information-dense hour-long videos. 
To overcome such limitations, we propose the \textbf{D}eep \textbf{V}ideo \textbf{D}iscovery (\textbf{DVD}) agent to leverage an \textit{agentic search} strategy over segmented video clips. 
Unlike previous video agents that rely on predefined workflows applied uniformly across different queries, our approach emphasizes the autonomous and adaptive nature of agents.
By providing a set of search-centric tools on multi-granular video database,
our DVD agent leverages the advanced reasoning capability of LLM to plan on its current observation state, strategically selects tools to orchestrate adaptive workflow for different queries in light of the gathered information.
We perform comprehensive evaluation on multiple long video understanding benchmarks that demonstrates our advantage. 
Our DVD agent achieves state-of-the-art performance on the challenging LVBench dataset, reaching an accuracy of \textbf{74.2\%}, which substantially surpasses all prior works, and further improves to \textbf{76.0\%} with transcripts.
The code has been released at \url{https://github.com/microsoft/DeepVideoDiscovery}.
\end{abstract}

\begin{figure*}[ht]
\vskip -0.1in
\begin{center}
\centerline{\includegraphics[width=\textwidth]{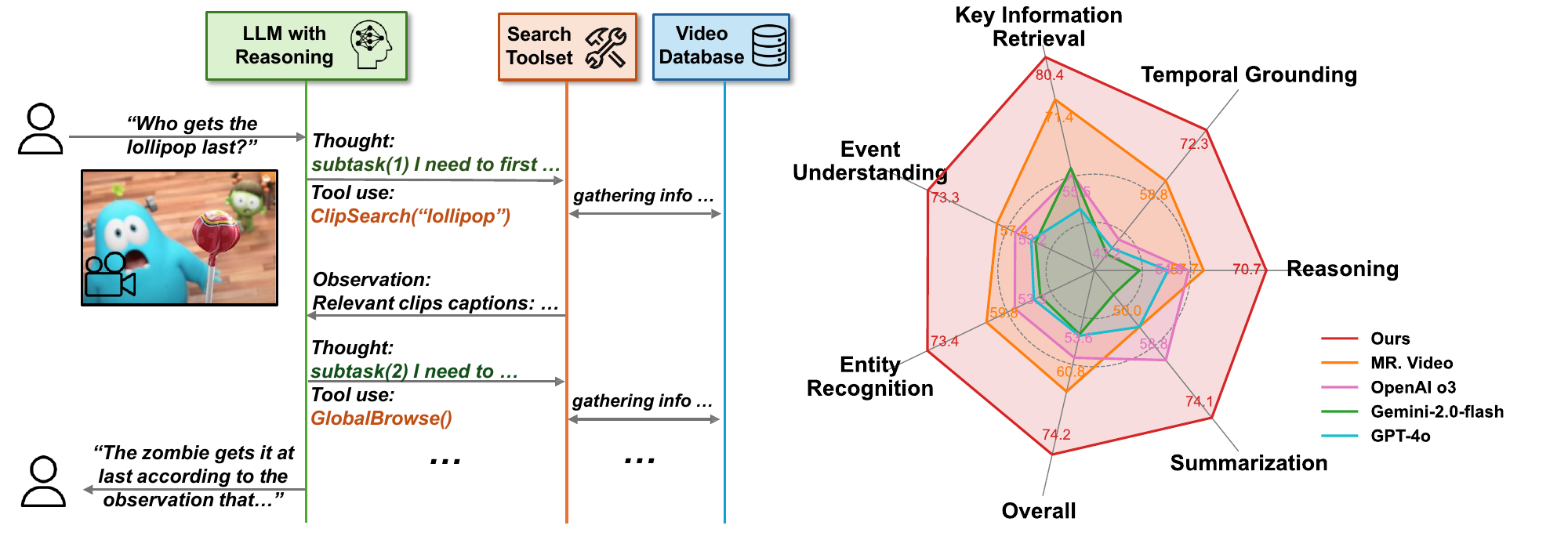}}
\caption{\textbf{Left:} Illustration of our Deep Video Discovery agent, which autonomously reasons on user query, iterative use tools to obtain the final answer. 
\textbf{Right:} Performance comparison on LVBench.}
\label{fig:gap_of_knowledge}
\end{center}
\end{figure*}
\vskip -0.1in

\section{Introduction}
\label{sec:intro}

Long-form videos are ubiquitous in everyday life, spanning diverse domains such as movies, meeting recordings, sports games, and variety shows. 
Accurately comprehending and interpreting content within these extensive videos remains an intrinsically challenging task~\cite{fu2024video,wang2024lvbench,wu2024longvideobench}, demanding an ability to simultaneously integrate and reason about intricate spatiotemporal details across broad global contexts. 
Effective retrieval of relevant information from hour-long or even longer sequences not only necessitates attending to fine-grained local details but also simultaneously interpreting subtle semantic relations distributed throughout extended temporal intervals.
Recent advancements in Large Language Models (LLMs) and Large Vision-language Models (VLMs) have notably improved capabilities in video understanding~\cite{openai2025gpt4.1,bai2025qwen2,wang2025adaretake} and increased context length handling more than one million tokens~\cite{openai2025gpt4.1,team2024gemini,yang2025qwen2}. 
However, even this extended context length remains insufficient for comprehending the information density typically found in long-form videos of hour-long duration. 
Empirical observations~\cite{openai2025gpt4.1} also suggest a decline in the model's effective instruction-following ability and reasoning clarity as the temporal dimension and information density increase.

Concurrently, recent breakthroughs~\cite{ds2025r1,openai2025o3} on reasoning capability of LLMs have facilitated advances in  agentic systems capable of complex information gathering tasks, such as Deep Research~\cite{openai2025deepresearch,gemini2025deepresearch,perplexity2025deepresearch} or Deep Search~\cite{jina2025deep,gork2025deep}. These agentic approaches demonstrate how decomposing difficult tasks into modular sub-tasks enables iterative reasoning, information searching, and content synthesis. Inspired by these successes, 
we conceptualize the challenge of understanding extremely long videos as a multi-step information search problem, where the video serves as the exploration environment and is segmented into multiple shorter video clips serving as manageable units of information. We name our approach as \ourmethod (Fig.~\ref{fig:gap_of_knowledge}, left).

While existing video agent frameworks~\cite{yang2024vca,fan2024videoagent,pang2025mr,wang2024videotree} incorporate searching processes in their designs, they manually design the search process with their human prior. For instance, both VideoTree~\cite{wang2024videotree} and VCA~\cite{yang2024vca} employ tree-based search strategies that navigate from root nodes to leaf nodes. This approach alleviates the context length limitations of VLMs but 
is inefficient for fine-grained queries which might benefit more from direct retrieving among leaf nodes. Additionally, semantically relevant entities may not exhibit temporal proximity, potentially diminishing the efficiency of backdate mechanism in tree-based search methods. 
Most critically, these existing frameworks prescribe fixed workflows that are uniformly applied across all query types, failing to adapt to the diverse information needs and optimal search strategies required by different questions.

In contrast to existing systems that typically rely on manually defined, rigid workflows, our approach is distinctly designed around an autonomous and adaptive agentic search paradigm for different queries. Instead of explicitly prescribing task workflows or search behaviors, we develop modular search tools that operate at multiple granularities, including (1) \textit{Global Browse}, (2) \textit{Clip Search}, and (3) \textit{Frame Inspect}. 
Global Browse enables global summarization and indexing of subjects and global contexts across the entire video. 
Clip Search implements efficient semantic retrieval of relevant events within segmented clips. Specifically, Frame Inspection empowers the agent to extract fine-grained details directly from pixel-level information in a specified temporal range. 

Equipped with this search-centric toolkit and multi-granular video database, our agent is inherently capable of autonomous reasoning, dynamic strategy orchestration, and iterative decision-making to proactively discover and extract crucial evidence. 
By leveraging the sophisticated reasoning capabilities intrinsic in the latest LLM, our agent does not merely use these tools independently, but adaptively combines their complementary strengths into a chain of thoughts and tool uses, effectively addressing diverse temporal-spatial and complex questions for long video. In the end, Deep Video Discovery can autonomously reason, plan, and retrieve pertinent information through adaptive search strategies tailored to specific video understanding queries.

We conduct comprehensive evaluations on long video benchmarks, demonstrating the efficiency and strong performance of our agent. In particular, on the challenging LVBench, we push forward the state-of-the-art performance by a large margin to 74.2\% (as shown in Fig.~\ref{fig:gap_of_knowledge}, right), further achieving 76.0\% with auxiliary transcripts.  We also set a series of ablation studies that show the effectiveness of our tool design. In addition, we analyze the behavior patterns of different reasoning models in tool use sequences, providing future insight of developing agents for long video understanding tasks.

\section{Related Work}

\paragraph{Long Video Understanding.}
Long video understanding remains a formidable challenge due to the intricate demands of temporal and spatial reasoning over extended durations and the complexity of information retrieval~\cite{wu2024longvideobench,wang2024lvbench}.
Recent efforts in VLM for long video understanding primarily tackle challenges of limited input frame number by extending the context length of models~\cite{team2024gemini,chen2024longvila} or minimizing video redundancy to reduce visual token numbers~\cite{han2024dynfocus,liu2025videoxlpro,wang2025adaretake}. Approaches such as \AdaReTaKe~\cite{wang2025adaretake} dynamically compress visual token by allocating adaptive compression ratios across time and model layers, thus significantly expanding the effective input frame number. 
However, token compression inherently introduces uncertainty regarding information loss, and models continue to face difficulties when answering complex queries under elongated context windows.
In parallel, given the sparsity of key information about the given query, some works~\cite{wang2024videoagent,fan2024videoagent,yang2024vca,wang2024videotree,pang2025mr,yan2025empowering} propose to explore the video content by agentic system. But they usually manually guide the agent about the search workflow by their priors~\cite{yang2024vca,wang2024videotree} or only allow the agent at simplex frame granularity to search~\cite{wang2024videoagent}, which cannot make full use of the reasoning capability of LLMs,  resulting in suboptimal search efficiency and a lack of holistic, global comprehension of the long video content.

\paragraph{Agent and tool use.}
Recent advancements in large language models (LLMs), particularly their enhanced reasoning and planning capabilities, have significantly accelerated the development of autonomous agents~\cite{yao2023react,zhang2023responsible,zhang2024codeagenttool,zhang2025ui}. The ability to leverage external tools~\cite{schick2023toolformer,qu2025tool,qin2024tool} further narrows the gap between general-purpose LLMs and real-world applications, enabling LLMs to acquire information, perform planning, and execute actions in complex environments. Our work extends this line of research to long video understanding, contributing to the broader investigation of solving complex video understanding tasks by integrating the advanced reasoning capabilities of LLMs with tool use. We introduce a suite of search-centric tools that allow LLMs to autonomously gather information at varying levels of granularity. By dynamically composing these tools, the agent can construct multi-step tool-use chains to improve the ability to answer complex queries effectively.

\begin{figure*}[t]
\begin{center}
\centerline{\includegraphics[width=\textwidth]{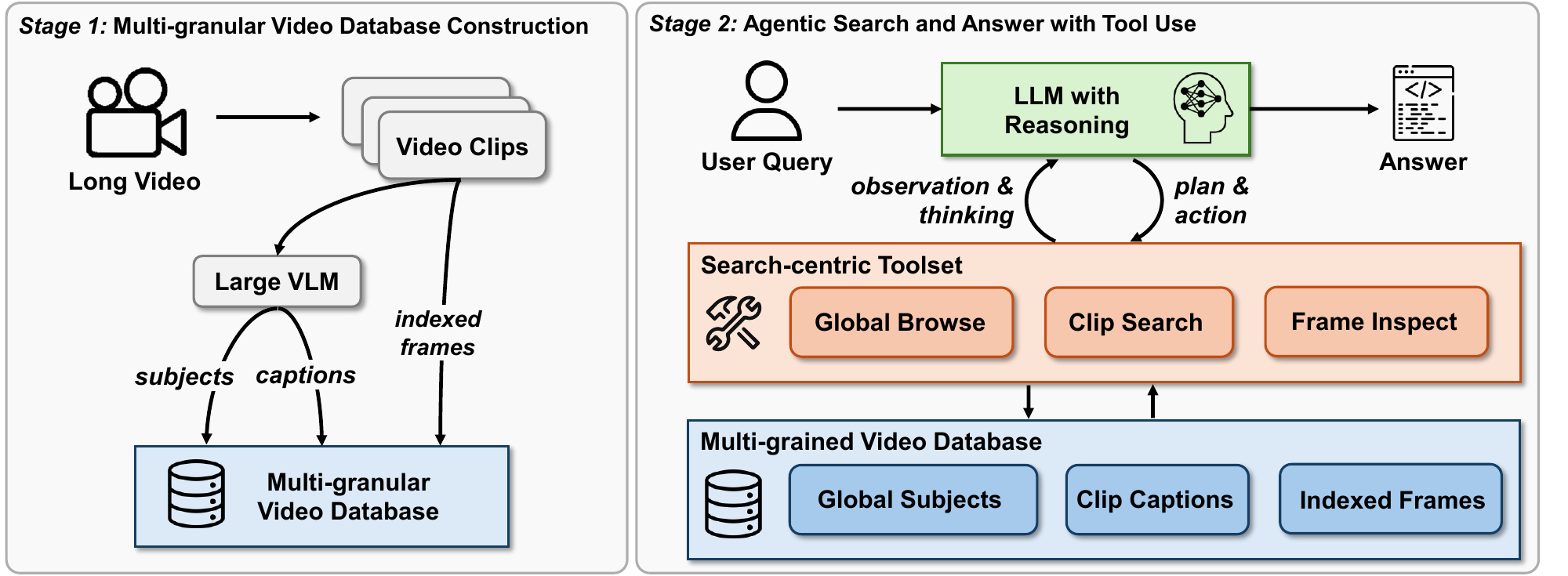}}
\caption{\ourmethod{}consists of two stages: 1) Multi-granular Video Database Construction. We extract video information from different levels to enable comprehensive understanding, efficient retrieval, and preservation of original content. 2) Agentic Search and Answer. The agent iteratively reasons on user query and leverage the tailored toolset to gather information to answer.}
\label{fig:pipeline}
\end{center}
\end{figure*}

\section{Deep Video Discovery}
\label{sec:method}




\paragraph{Overview.} 
To solve the long-form video understanding problem in an agentic search way,
we first build a multi-grained structured database from the long-form video. The database then serves for search-centric tools that work at different granularities.
Specifically, our Deep Video Discovery agent consists of three main components: the multi-grained video database $\mathcal{D}$, search-centric toolset $\mathcal{T}$, and the LLM \( M \) as the agent's orchestrator. 
Given the user query $Q$, the agent reasons iteratively to choose an action $A_i \in \mathcal{T}\cup\{\textsc{Answer}\}$ with parameters $P$ to gather information for the video database $\mathcal{D}$ or make decision to answer the query by referring to the accumulated information in this process.
In the following subsections, we sequentially introduce the multi-grained video database construction and Agentic Search and Answer with Tool Use.






\subsection{Multi-granular Video Database Construction} 
\label{subsec:database}

Given an ultra long input video $V$, our goal is to transform it into a database that can provide efficient fast retrieval and also provide the original pixels of video for detailed information when necessary.
Hence, we design it in a multiple granularity style which can provide different levels of video information for corresponding search tools.
Specifically, we first segment the video into clips as the basic information unit then make the
database include global summarized information to cover the whole video, a clip-based caption corpus and indexed frames from the clip. Fig.~\ref{fig:pipeline} (left) provides an overview.
We introduce these components sequentially.


\paragraph{Temporal segmentation.}  

We start by uniformly partitioning the input video $V$ into a temporal sequence of non-overlapping short clips $\{v_i\}_{i=1}^{N}$, where the total segments $N=\lceil\frac{\text{len}(V)}{t}\rceil $. 
Empirically, we set $t=5$ seconds to provide an adequate balance between computing cost and semantic and action completeness. 
Then all the video clips are decoded into frames $\{f_i\}_{i=1}^{N}$ under 2 frames per second for further process.

\paragraph{Multi-granular information extraction.} 
\label{sec:database}


Our multi-granular video information is designed as three levels: global video level, clip level and frame level. Specifically, at the global level we summarize the video content into a compact, subject-centric representation. At the clip level, we leverage textual captions to facilitate efficient information retrieval, while at the frame level we preserve original decoded frames indexed according to their corresponding clips, enabling precise reference and detailed analysis when required.

To derive the subject-centric global representation while minimizing redundancy in caption generation, we maintain an progressive structured subject registry $S$ throughout the clip captioning process.
Specifically, given a video clip $v_i$ and decoded frame $f_i$, we prompt a large VLM to generate the captioning $c_i$ and evolve registry whenever new subjects appear. The process is denoted as $S_i, c_i=VLM(f_i, S_{i-1})$ where $S_0$ is initialized as empty, and at the conclusion of the captioning process, the final subject registry is denoted by $S=S_N$. Each subject within the registry is represented by a comprehensive set of attributes, including name, physical appearance, identity descriptors, associated actions, and corresponding temporal spans in the video.
The obtained caption $c_i$ is subsequently embedded into a dense semantic vector $e_i\in\mathbb{R}^d$ using a language embedding model, facilitating fast retrieval in downstream applications.
Despite careful design choices, perceptual compression inherent in caption generation inevitably entails some information loss. To mitigate this when necessary, we explicitly retain the decoded frames $f_i$ alongside their corresponding textual captions and embeddings.

\paragraph{Outcome.} 
 The finalized database therefore encapsulates the decoded frames, captions and corresponding embedding triples, thus forming a structured database
$
    \mathcal{D} = \{S, \{{f_i,\, c_i,\, e_i\}}_{i=1}^{N}\}.
$
This offline construction procedure transforms a lengthy raw video into a structured set of textually searchable embeddings with associated clips, while simultaneously preserving the complete visual content at pixel resolution. The resulting database becomes the basis for adaptive tool usage, enabling global information browsing, efficient semantic retrieval at the video-clip scale, and comprehensive grounding of generated outputs back to their source frames.

\begin{table}[t]
    \centering
    \begin{minipage}[t]{0.42\textwidth}
        \centering
        \label{tab:actions_no_return} 
        \caption{Action space overview of our DVD. The first three actions are from our toolset and the final \textsc{Answer} action is designed as stop criterion.} 
        \vspace{1mm}
        \resizebox{\textwidth}{!}{
        \begin{tabular}{ll} 
            \toprule 
            \textbf{Action} & \textbf{Parameter} \\ 
            \midrule 
            \multirow{2}{*}{\centering \textsc{\shortstack{Global \\ Browse}}} & video database $\mathcal{D}$ \\
            & user query ${Q}$ \\
            \midrule 
            \multirow{3}{*}{\centering \textsc{\shortstack{Clip \\ Search}}} & video database $\mathcal{D}$ \\
            & agent synthesized query $\hat{Q}$ \\
            & return top-$k$ captions\\
            \midrule 
            \multirow{3}{*}{\centering \textsc{\shortstack{Frame \\ Inspect}}} & video database $\mathcal{D}$ \\
            & agent synthesized query $\hat{Q}$ \\
            & temporal range $[t_s, t_e]$ \\
            \midrule 
            \textsc{Answer} & the answer to user query \\
            \bottomrule 
        \end{tabular}
        }

    \end{minipage}%
    \hfill 
    \begin{minipage}[t]{0.52\textwidth}
        \centering
        \small
        \begin{algorithm}[H]
            \caption{Agentic Search and Answer. 
            }
            \label{algo:asa}
            \SetKwInOut{Input}{Input}
            \SetKwInOut{Output}{Output}
            \Input{Initial query $Q$, max step $N$, LLM $M$, tool set $\mathcal{T}$, action space $\mathcal{A}=\mathcal{T} \cup \{\textsc{Answer}\}$}
            \Output{Answer to $Q$}
            Initialize history $H_0 \leftarrow \{Q, \mathcal{A}\}$ \\
            \For{$i \leftarrow 1$ \KwTo $N$}{
                $R_i \leftarrow M.reason(H_{i-1})$ \\
                $A_i, P_i \leftarrow M.call(R_i, H_{i-1})~where~A_i\in \mathcal{A}$ \\
                \If{$A_i = \textsc{Answer}$}{ 
                    \textbf{break}\\ 
                }
                $O_i \leftarrow A_i(P_i)$  \\ 
                $H_i \leftarrow H_{i-1} \cup \{(R_i, A_i, O_i)\}$ \\
                \If{$t=N$}{
                    $P_i \leftarrow M.answer(H_i)$ \\
                }
            }
            \Return{$\textsc{Answer}(P_i)$}
        \end{algorithm}
    \end{minipage}
\end{table}

\subsection{Agentic Search and Answer with Tool Use}
With the built multi-granular video database, we design a set of search-centric tools that can enable global information understanding, efficient clip retrieval by semantic query, and details exploration on original video content.
By equipping a reasoning large language model with this toolset, we build our DVD that can address complex user query on long video though autonomous planning and strategical search tool combination, as shown in Fig.~\ref{fig:pipeline} (right).
We refer to this stage as \textbf{Agentic Search and Answer with Tool Use} (\textbf{ASA}).
We introduce this stage through two subsections: Search-centric Tool Preparation and Agent Design.


\subsubsection{Search-centric Tool Preparation} 
\label{subsec:tool}

Leveraging the established video database, we have developed a suite of tools designed to efficiently gather information from video data at varying levels of granularity. 
Specifically, we divide long videos into three distinct hierarchical levels and introduce corresponding specialized tools: (1) Global Browse, (2) Clip Search, and (3) Frame Inspect.
Given the significant computational cost associated with processing lengthy videos using VLMs, our tool design carefully balances efficiency and performance. Central to our approach is an agentic search paradigm, wherein the agent decomposes the user query and strategically chains up tools with synthesized parameters, enabling iterative reasoning and information collection to resolve the task.
Through the effective integration and coordinated use of these tools, the agent progressively enhances its understanding of user intent and precisely locates relevant information within extensive video content.
We introduce the three tools sequentially in the following paragraphs.

\paragraph{Tool: Global Browse.}
\label{tool:global}
The Global Browse tool takes the video database and the original user query  as input, and returns global summaries capturing high-level contextual information.
We construct two distinct types of global information: subject-centric and event-centric summaries. 
For subject-centric summarization, we pre-construct it when building the multi-granular video dataset as mentioned in Section~\ref{sec:database} since it is query-irrelevant. 
For event-centric summarization, we uniformly sample frames across the entire video and feed these sampled frames into the VLM. We instruct the VLM to describe noteworthy events explicitly related to the original user query. 
Upon invocation by the agent, the Global Browse tool efficiently retrieves and returns these global representations, providing the agent immediate access to high-level global context information.

\paragraph{Tool: Clip Search.}
\label{tool:clip}
Clip Search provides a mid-level granularity retrieval capability, enabling fast and efficient exploration of video content via caption embedding. Given a query $\hat{Q}$ synthesized based on the agent’s current internal reasoning context, this module retrieves a ranked list of top-$k$ relevant video clips along with their captions. 
Specifically, the tool computes the cosine similarity between the embedding of the provided query and the pre-computed embeddings of all video clip captions, returning the clips corresponding to the highest-ranked caption matches. 
Each retrieved observation contains both the corresponding caption and the time ranges of the associated video clip. 
To achieve an accurate and detailed understanding, the agent can iteratively invoke this tool, progressively refining temporal constraints or reformulating its queries based on newly acquired contextual knowledge. This iterative chain-of-query approach effectively guides the agent toward precise temporal segments relevant to the original high-level query.

\paragraph{Tool: Frame Inspect.}
\label{tool:frame}
Frame Inspect receives a temporal range $[t_s, t_e]$ within the video and an sub-query freely defined by the agent as input, returning an open format visual-question-answering (VQA) response. 
The agent can invoke this tool whenever explicit frame-level details such as subtle attributes, object counting, or fine-grained spatial relationships, are required but not clearly depicted in captions or global summaries. The open-ended query format allows significant freedom for the agent to leverage its reasoning capability, enabling highly adaptable visual inspection. 
Specifically, the Frame Inspect tool loads raw frames from the requested interval and prompts a VLM with these frames and agent-synthesized query. 
To ensure computational efficiency, we limit processing to a maximum of 50 frames, uniformly sampling from frames exceeding this limit. The resulting response thus equips the agent with accurate, visually-grounded evidence essential for detailed reasoning tasks.

\subsubsection{Agentic Design}
\label{sec:asa}


To maximally leverage the reasoning and planning capacity intrinsic to modern LLMs, we intentionally abstain from manually instructing explicit seaching workflow or tool utilization patterns.
Instead, we enable the agent to reason, plan, and take actions through a streamlined iterative \textit{observe-reason-act} loop, similar to ReAct\cite{yao2023react}. 
For a given query, the agent reasons about its current observation state, strategically selects search tools, formulates appropriate parameters for actions, and dynamically refines its internal reasoning in light of the gathered evidence.
Within ASA, the LLM acts as a sophisticated cognitive driver, taking actions at each iteration based on cumulative knowledge and reasoned evidence, thereby reinforcing its pivotal role in adaptively navigating the discovery process.



Specifically, as illustrated in Algorithm~\ref{algo:asa}, given an initial user query \( Q \), a predefined action space \( \mathcal{A} = \mathcal{T} \cup \{\textsc{Answer}\} \), and a maximum allowable step count \( N \), our agent performs iterative reasoning to strategically navigate the available actions. The agent leverages an LLM \( M \) to reason upon the current dialogue history, plan its immediate action, interact with the toolset \( \mathcal{T} = \{\textsc{GlobalBrowse}, \textsc{ClipSearch}, \textsc{FrameInspect}\} \), and collect observations \( O_{i} \). More concretely, at each step \( t \), the agent maintains a historical context \( H_{i} \), reflects to generate a reasoning step \( R_{i} \), selects an action \( A_{i} \in \mathcal{T}\cup\{\textsc{Answer}\} \) accompanied by relevant parameters \( P_{i} \), and receives subsequent observation outcomes \( O_{i} \) from the environment. These components, reasoning, action, and obtained outcomes, are successively appended to the interaction history \( H_{i} \), enriching the context for subsequent iterations of inference. The iterative process terminates either when the agent explicitly selects the \textsc{Answer} action, or upon reaching the step limit \( N \), at which prompts the agent directly generates a final answer prediction. The agent then outputs the final answer to the original user query. 

By positioning the LLM's sophisticated reasoning at the core of this iterative loop, this approach endows the agent with an inherently autonomous, evidence-guided, and flexible action-taking mechanism.
This autonomous and iterative paradigm fosters a strategic and context-sensitive inquiry cycle, thereby enabling the agent to effectively leverage the available tools to iteratively decompose the original query into progressively refined sub-queries, updating and improving the query representation as it receives new observations. Through iterative reasoning and interaction cycles, guided by deeper and increasingly comprehensive observations collected from prior tool usage, the agent systematically enhances its understanding and interpretation of the task context, ultimately leading to more accurate and informed answers to the given question.




\section{Experiment}
\label{sec:exp}

\begin{table}[t]
    \caption{Comparison on LVBench under different question categories. The metric shown is accuracy (\%).}
    \centering
    \renewcommand\arraystretch{1.15}
    \resizebox{0.88\linewidth}{!}{
    \begin{threeparttable}
        \begin{tabular}{l cccccc c}
            \toprule
            Methods & ER & EU & KIR & TG & Rea & Sum & Overall \\
            \midrule
            \multicolumn{8}{l}{\textcolor{gray}{\textit{Commercial VLMs}}} \vspace{0.2mm} \\
            Gemini-1.5-Pro~\cite{team2023gemini} & 32.1 & 30.9 & 39.3 & 31.8 & 27.0 & 32.8 & 33.1  \\
            Gemini-2.0-Flash~\cite{team2023gemini} & 47.4 & 48.5 & 56.8 & 39.3 & 44.4 & 41.4 & 48.6  \\
            GLM-4V-Plus~\cite{glm2024chatglm} & 46.2 & 47.8 & 54.1 & 42.7 & 46.5 & 37.9 & 48.7  \\
            GPT-4o~\cite{achiam2023gpt} & 48.9 & 49.5 & 48.1 & 40.9 & 50.3 & 50.0 & 48.9  \\
            OpenAI o3~\cite{openai2025o3} & 57.6 & 56.4 & 62.9 & 46.8 & 50.8 & 67.2 & 57.1  \\
            \midrule
            \multicolumn{8}{l}{\textcolor{gray}{\textit{Open-Source VLMs}}} \vspace{0.2mm} \\
            InternVL2.5-78B~\cite{wang2025internvideo2} & 43.8 & 42.0 & 42.1 & 36.8 & 51.0 & 37.9 & 43.6  \\
            VideoLLaMA3-7B~\cite{damonlpsg2025videollama3} & 45.8 & 42.4 & 47.8 & 35.9 & 45.8 & 36.2 & 45.3  \\
            Qwen2.5-VL-72B~\cite{bai2025qwen2} & - & - & - & - & - & - & 47.7  \\
            VideoChat-Flash~\cite{li2024videochat} & 51.1 & 46.0 & 49.0 & 38.9 & 48.5 & 34.5 & 48.2 \\
            \AdaReTaKe~\cite{wang2025adaretake} & 53.0 & 50.7 & 62.2 & 45.5 & 54.7 & 37.9 & 53.3  \\
            \midrule
            \multicolumn{8}{l}{\textcolor{gray}{\textit{Video Agents and Others}}} \vspace{0.2mm} \\
            VideoTree~\cite{wang2024videotree} & 30.3 & 25.1 & 26.5 & 27.7 & 31.9 & 25.5 & 28.8  \\
            VideoAgent~\cite{wang2024videoagent} & 28.0 & 30.3 & 28.0 & 29.3 & 28.0 & 36.4 & 29.3  \\
            VCA~\cite{yang2024vca} & 43.7 & 40.7 & 37.8 & 38.0 & 46.2 & 27.3 & 41.3  \\
            MR. Video~\cite{pang2025mr} & 59.8 & 57.4 & 71.4 & 58.8 & 57.7 & 50.0 & 60.8  \\
            \midrule
            \textbf{Deep Video Discovery (Ours)} & \textbf{73.4} & \textbf{73.3} & \textbf{80.4} & \textbf{72.3} & \textbf{70.7} & \textbf{74.1} & \textbf{74.2} \\
            {\textcolor{black}{\ + Auxiliary transcripts}} & \textbf{\textcolor{black}{75.5}} & \textbf{\textcolor{black}{77.1}} & \textbf{\textcolor{black}{79.0}} & \textbf{\textcolor{black}{72.7}} & \textbf{\textcolor{black}{68.7}} & \textbf{\textcolor{black}{84.5}} & \textbf{\textcolor{black}{76.0}} \\
            \bottomrule
        \end{tabular}
    \end{threeparttable}
    }
    \label{tab:lvbench_comparison}
\end{table}

\subsection{Benchmarks}

We assess the long-form video understanding capabilities of Deep Video Discovery using several established long video benchmarks. Our primary evaluation benchmark, \textbf{LVBench}~\cite{wang2024lvbench}, includes 1,549 multiple-choice questions across 103 hour-long videos. 
It stands as one of the most comprehensive and challenging benchmarks for extreme long-form video understanding. 
\textbf{LongVideoBench}~\cite{wu2024longvideobench} features 6,678 questions from 3,763 videos, ranging in duration from a few seconds to an hour. 
We emphasize the longest subset with durations in $(900s, 3600s]$ (denoted as the \textit{Long} subset), comprising 564 questions from 188 videos. 
\textbf{Video MME}~\cite{fu2024video} is segmented by video duration; we concentrate on the \textit{Long} subset without subtitles to isolate long-video comprehension, covering 300 videos of 30 to 60 minutes with 900 questions. 
Finally, \textbf{EgoSchema}~\cite{mangalam2023egoschema} serves as a diagnostic benchmark for long-video understanding, 
where we evaluate on its validation split of 500 videos with 500 questions.

\subsection{Implementation Details}
\label{sec:details}
\textbf{Baselines}. We compare Deep Video Discovery with a range of long-video understanding systems, including both VLM-based~\cite{team2023gemini, achiam2023gpt, openai2025o3, glm2024chatglm, ye2024mplug, wang2025internvideo2, damonlpsg2025videollama3, bai2025qwen2, li2024videochat, wang2025adaretake} and agent-based approaches~\cite{wang2024videotree, fan2024videoagent, yang2024vca, pang2025mr}. Most baseline results are taken from official leaderboards or published reports, except for the recently released OpenAI o3~\cite{openai2025o3}, which has not yet been evaluated on these benchmarks. Following \cite{pang2025mr}, we uniformly sample 256 frames per video to evaluate OpenAI o3.

\textbf{Deep Video Discovery} flexibly integrates different models depending on the needs of each component. For the VLM in video database construction, we use GPT-4.1~\cite{openai2025gpt4.1} to produce high-quality captions on LVBench, and GPT-4.1-mini for other benchmarks to reduce cost. During agentic search and answering, we employ OpenAI o3 as LLM \( M \) for its strong reasoning ability, including in the Frame Inspect module for fine-grained VQA. All frames are resized to 720p to maintain visual details. 
In Clip Search, we set 16 as the default value of top-$k$ while leaving the flexibility for LLM to change it. 
Maximum reasoning step is set to $N=15$. To explore the upper bound of understanding ability, we additionally evaluate LVBench using \textit{auxiliary transcripts}. Audio is transcribed with WhisperX\cite{bain2022whisperx}, and transcripts are used to guide video segmentation and enrich captions. This audio-visual fusion enhances understanding of long, complex content, leading to stronger results.

\textbf{API Content filtering.}
We use LLM API via Azure OpenAI Service. We observe that the safety content filtering mechanism of the service misjudges a small part of data from the benchmark as offensive and block the request, which leads to the reduced performance of both OpenAI o3 baseline and our DVD agent. We provide more details and mitigation strategies in Appendix~\ref{sec:api}.

\begin{table}[t]
    \caption{Comparison on long video benchmarks. }
    \centering
    \renewcommand\arraystretch{1.15}
    \resizebox{0.99\linewidth}{!}{
    \begin{threeparttable}
        \begin{tabular}{l ccccc}
            \toprule
            \multirow{2}{*}{Methods}   & LVBench & \multicolumn{2}{c}{LongVideoBench (Val)}   & Video MME    & EgoSchema \\
            ~       & Overall & Overall & Long  & Long (w/o sub)  & Val \\
            \midrule
            \multicolumn{4}{l}{\textcolor{gray}{\textit{Commercial VLMs}}} \vspace{0.2mm} \\
            Gemini-1.5-Pro~\cite{team2023gemini}  & 33.1 & 64.0 & 58.6 & \textbf{67.4}  & -  \\
            Gemini-2.0-Flash~\cite{team2023gemini}  & 48.3 & - & 45.7 & 63.0  & 71.2  \\
            GPT-4o~\cite{achiam2023gpt} & 48.9  & 66.7 & 60.9 & 65.3  & 70.4  \\
            OpenAI o3~\cite{openai2025o3} & 57.1  & 67.5 & 60.6 & 64.7 & 63.2 \\
            \midrule
            \multicolumn{4}{l}{\textcolor{gray}{\textit{Open-Source VLMs}}} \vspace{0.2mm} \\
            mPLUG-Owl3~\cite{ye2024mplug} & 43.5  & 59.8 & - & 50.1  & -  \\
            InternVL2.5-78B~\cite{wang2025internvideo2} & 43.6  & 63.6 & - & 62.6 & -  \\
            Qwen2.5-VL-72B~\cite{bai2025qwen2} & 47.7  & 60.7 & - & 63.9  & -  \\
            \AdaReTaKe~\cite{wang2025adaretake} & 53.3  & 67.0 & - & 65.0 & -  \\
            \midrule
            \multicolumn{4}{l}{\textcolor{gray}{\textit{Video Agents and Others}}} \vspace{0.2mm} \\
            VideoTree~\cite{wang2024videotree} & 28.8 &  - & - & -  & 67.0  \\
            VideoAgent~\cite{wang2024videoagent} & 29.3  & - & - & - & 63.2  \\
            VCA~\cite{yang2024vca} & 41.3 & - & - & -  & 73.6  \\
            MR. Video~\cite{pang2025mr} & 60.8 & - & 61.6 & 61.8 & 73.0  \\
            \midrule
            \textbf{Deep Video Discovery (Ours)} &\textbf{74.2} &  \textbf{71.6} & \textbf{68.6}  & 67.3 & \textbf{76.6}  \\
            \bottomrule
        \end{tabular}
    \end{threeparttable}
    }
    \label{tab:long_video_benchmarks}
\end{table}

\subsection{Main Results}

Table~\ref{tab:lvbench_comparison} presents the comparison results on LVBench. DVD significantly outperforms all baselines, surpassing the previous SOTA MR. Video by 13.4\%. Compared to the prior leading video agent VCA, our method achieves a remarkable 32.9\% gain. Against our base VLM, OpenAI o3, our full system delivers a substantial 17.1\% gain, highlighting the importance of agentic reasoning. Incorporating transcript information provides an additional 1.8\% boost. These results highlight the effectiveness of our search-centric agentic reasoning framework in handling ultra-long video understanding tasks.

Table~\ref{tab:long_video_benchmarks} provides a comprehensive evaluation across several long-video benchmarks. On LongVideoBench, DVD outperforms the previous SOTA by 4.1\% overall and 7.0\% on the longest-duration subset. On the Video MME Long subset, it beats the best open-source VLM, \AdaReTaKe, by 2.3\%, and MR. Video by 5.5\%, approaching the performance of Gemini-1.5-Pro. On EgoSchema, our method exceeds the previous best by 3.0\%. Notably, it exceeds reported human-level accuracy of $\sim$76\% on this benchmark. Across all datasets, our system consistently outperforms the base VLM OpenAI o3, confirming the effectiveness and generalizability of our agentic reasoning framework.

\begin{table}[t]
\begin{minipage}[t]{0.51\textwidth}
    \caption{\small Ablation on used models. $M_{database}$ for captioning in database construction, $M_{reasoning}$ for reasoning in ASA, $M_{tool}$ for Frame Inspect.}
    \centering
    \renewcommand\arraystretch{1.15}
    \resizebox{\linewidth}{!}{
    \begin{threeparttable}
        \begin{tabular}{ccc|c}
        \toprule
        \multicolumn{3}{c|}{Adopted models} & LVBench \\
        $M_{database}$  & $M_{reasoning}$ & $M_{tool}$  & w/ transcripts \\\midrule
        4.1         & o3        & 4.1-mini  & 72.3 \\ 
        4.1         & o4-mini   & o3        & 70.2 \\
        4.1         & 4o        & o3        & 62.3 \\
        4.1-mini    & o3        & o3        & 71.9 \\ \midrule 
        4.1         & o3        & o3        & \textbf{76.0}  \\ 
        \bottomrule
        \end{tabular}
    \end{threeparttable}
    }
    \label{tab:ablation-vlms}
\end{minipage}
\hfill
\begin{minipage}[t]{0.44\textwidth}
    \caption{\small Ablation on the search-centeric tools \( \mathcal{T} \). Note that the anchor uses 4.1-mini for $M_{database}$, and o3 for both $M_{reasoning}$ and $M_{tool}$.}
    \centering
    \renewcommand\arraystretch{1.15}
    \resizebox{\linewidth}{!}{
    \begin{threeparttable}
        \begin{tabular}{ccc|c}
        \toprule
        \multicolumn{3}{c|}{Search-centeric Tools} & LVBench \\ 
        \makecell{Global\\Browse} & 
    \makecell{Clip\\Search} & 
    \makecell{Frame\\Inspect} & w/ transcripts \\ \midrule
                    & \checkmark    & \checkmark    &  69.0  \\
        \checkmark  &               & \checkmark    &  59.6 \\
        \checkmark  & \checkmark    &               &  63.5 \\
        \midrule
        \textbf{\checkmark}  & \checkmark    & \checkmark    &  \textbf{71.9} \\
        \bottomrule
        \end{tabular}
    \end{threeparttable}
    }
    \label{tab:ablation-tools}
\end{minipage}

\end{table}

\subsection{Ablation Study}

We evaluate the impact of different model choices across system components. 
By default, GPT-4.1 is used for captioning and subject extraction during Multi-granular Video Database Construction and OpenAI o3 serves as the reasoning model in the Agentic search and Answer with tool use process while Frame Inspect tool also leverages OpenAI o3 to query the fine-grained details on the frame pixels. 
We denote the three models as $M_{database}$, $M_{reasoning}$ and $M_{tool}$ in Table~\ref{tab:ablation-vlms}.
Replacing GPT-4.1 with GPT-4.1-mini for database construction or Frame Inspect tool results in moderate drops of 4.1\% and 3.7\%, respectively, indicating relatively minor impact. 
For reasoning model in agentic search, switching to OpenAI o4-mini~\cite{openai2025o3} leads to a 5.8\% drop, while GPT-4o causes a substantial 13.7\% decline. 
It highlights the reasoning model as the most critical component in our agentic system because our system is designed surrounding and to make full use of the reasoning capability of LLM. The lack of reasoning ability leads to the collapse of agent behavior, as analyzed further in the subsequent subsection.

We next examine how the choice of reasoning model affects performance when pairing DVD with various open-source LLMs on LVBench. As reported in Table~\ref{tab:opensourcellm}, combining DVD with the open-source DeepSeek-R1 reasoning model yields 68.5\% accuracy, surpassing all prior methods. Notably, even when restricted to a 32B model Qwen3-32B, DVD attains 57.3\% accuracy, outperforming existing VLM approaches, including GPT-4o and OpenAI o3.

Finally we assess the contribution of each tool in the agentic search and answer phase (Table~\ref{tab:ablation-tools}). Removing Global Browse which is responsible for global summarization and long-range event linking leads to a 2.9\% drop. Disabling Frame Inspect with the fine-grained VQA results in a 8.4\% decline, highlighting its role in fine-grained understanding. Removing Clip Search causes the largest drop of 12.3\%, as it breaks the searching ability for iteratively refine reasoning. These results underscore the importance of tool integration in our search-centric framework.

\begin{table}[t]
\begin{minipage}[t]{0.52\textwidth}
    \caption{Comparison of DVD agent with different LLMs as reasoning models on \textbf{LVBench}. * denotes Qwen3-32B-Thinking model.}
    \centering
    \renewcommand\arraystretch{1.15}
    \resizebox{\linewidth}{!}{
    \begin{threeparttable}
          \begin{tabular}{ccc}
    \toprule
    \textbf{Category} & \textbf{Model} & \textbf{Accuracy (\%)} \\
    \midrule
    \multirow{2}{*}{\shortstack{DVD w/\\closed-sourced}}
      & OpenAI o3          & \textbf{76.0 }\\
      & GPT-4o             & 62.3 \\
    \midrule
    \multirow{3}{*}{\shortstack{DVD w/\\open-sourced}}
      & DeepSeek-R1        & \textbf{68.5} \\
      & DeepSeek-V3        & 57.5 \\
      & Qwen3-32B$^*$      & 57.3 \\
    \bottomrule
  \end{tabular}
    \end{threeparttable}
    }
    \label{tab:opensourcellm}
\end{minipage}
\hfill
\begin{minipage}[t]{0.44\textwidth}
    \caption{\small Comparison DVD and its variation on different max step number.}
    \centering
    \renewcommand\arraystretch{1.15}
    \resizebox{\linewidth}{!}{
    \begin{threeparttable}
        \begin{tabular}{lccc}
\toprule
\textbf{Max Step Limit} & \textbf{8} & \textbf{12} & \textbf{15} \\
\midrule
\multicolumn{4}{l}{\textbf{Ours}} \\
\quad Avg.\ Actual Steps      & 6.7 & 7.2 & 7.3 \\
\quad Accuracy (\%)    & 72.3 & 73.8 & 74.2 \\
\midrule
\multicolumn{4}{l}{\textbf{Ours w/ VideoAgent~\cite{wang2024videoagent} workflow}} \\
\quad Avg.\ Actual Steps      & 5.0 & 8.3 & 11.1 \\
\quad Accuracy (\%)    & 48.4 & 66.3 & 70.2 \\
\bottomrule
\end{tabular}
    \end{threeparttable}
    }
    \label{tab:efficiency}
\end{minipage}

\end{table}

\begin{figure*}[t]
\begin{center}
\centerline{\includegraphics[width=\textwidth]{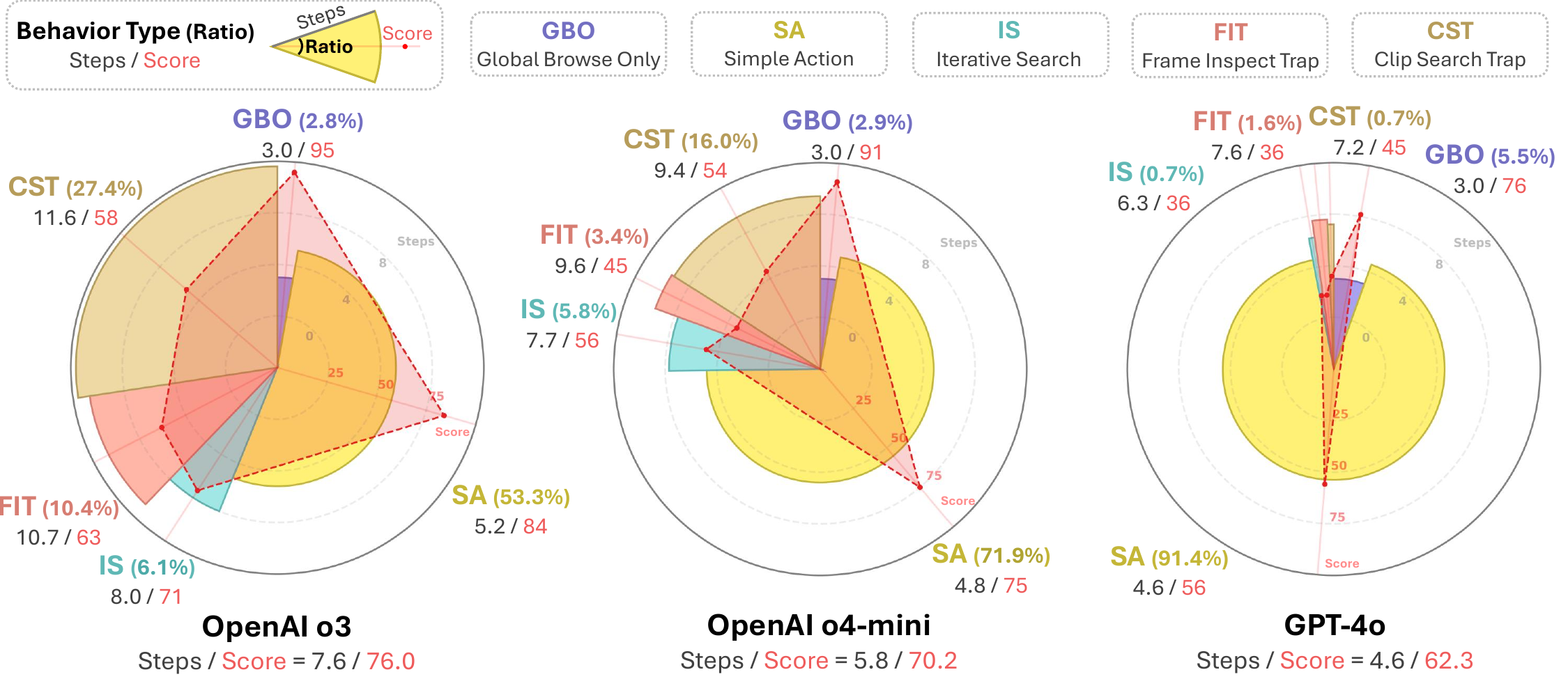}}
\caption{Analysis of the behavior of Deep Video Discovery using different \textit{reasoning} models. We categorize tool-calling behavior into five types. For each type, we report its proportion (\textit{Ratio}, sector angels), average reasoning steps (\textit{Steps}, sector radius) and score (\textit{Score}, dashed lines). A clear correlation emerges among behavior patterns, reasoning depth, and score (see Section~\ref{sec:exp-analysis} for details).}
\label{fig:analysis}
\end{center}
\end{figure*}

\subsection{Analysis on Efficiency}

To understand the efficiency of our designed agent, we compare DVD's performance with different maximum step limits on LVBench without transcripts. Since the core strength of DVD lies in its adaptive workflow orchestrated by the reasoning model, we also evaluate a variant that replaces this adaptive workflow with the hand-crafted workflow from VideoAgent~\cite{wang2024videoagent} while keeping all other settings identical.
Specifically, the VideoAgent workflow follows a fixed sequence: first predicting answers and confidence scores, then identifying missing information through self-reflection, searching video clips for the missing information, merging found information into current observations, and finally re-predicting answers and confidence scores.
As shown in Table~\ref{tab:efficiency}, the hand-crafted approach results in a significant 7.5\% absolute accuracy drop while requiring more steps. Even when scaled to 11.1 average steps (52\% more than our DVD), the hand-crafted workflow achieves only 70.2\% accuracy, still 4.0\% below our adaptive approach. This comparison highlights the superior search efficiency of our adaptive workflow design.

\subsection{Analysis on Agentic Reasoning Behavior}
\label{sec:exp-analysis}

The reasoning model is the most critical component in DVD. During the \textit{observe-reason-act} loop, the agent autonomously integrates current context and flexibly decides the next tool to invoke. To better understand this, we analyze the tool-calling behavior during the agentic search and answer phase and category it into five types for analysis (see Fig.~\ref{fig:analysis}). At the same time, we provide more detailed qualitative case study in Appendix~\ref{sec:casestudy}.

\textbf{Global Browse Only.} The agent confidently answers after a single Global Browse call. This pattern, though rare, correlates with very high accuracy, suggesting the initial global context was sufficient.

\textbf{Simple Action.} The agent follows a direct \textit{search-inspect-answer} path, typically involving one or two Clip Search and Frame Inspect calls. This is the most common and effective strategy, employed in over half of the queries and achieving high accuracy.

\textbf{Iterative Search.} The agent alternates between Clip Search and Frame Inspect multiple times. This indicates an initial difficulty in locating sufficient information, requiring a more exploratory approach. These trajectories are longer (\textit{e.g.}, 8.0 vs. 5.2 steps for OpenAI o3) and yield slightly lower accuracy than a Simple Action.

\textbf{Frame Inspect Trap.} The agent becomes stuck in a loop of fine-grained analysis, invoking Frame Inspect more than three consecutive times without reaching a conclusion. This behavior leads to prolonged reasoning chains and significantly lower accuracy.

\textbf{Clip Search Trap.} The agent repeatedly calls Clip Search more than three times consecutively, often when key information is absent from the video database. The agent fails to adapt, resulting in unproductive loops. This pattern is a primary failure mode for the OpenAI o3 model.

This analysis yields two key insights into designing effective autonomous video agents:

\textbf{Insight 1: The dual nature of reasoning length.} For a single model, longer reasoning trajectories often signal uncertainty and correlate with lower accuracy (\textit{e.g.}, Simple Action vs. Iterative Search). However, when comparing across models, superior performance is often achieved by the model capable of conducting more thorough, and thus longer, reasoning processes.

\textbf{Insight 2: Overconfidence leads to behavioral collapse.} The GPT-4o model significantly underperforms, collapsing into the Simple Action pattern for 91.4\% of queries. With an average of only 4.6 reasoning steps, it frequently concludes prematurely, rarely exploring alternative strategies. This ''overconfidence`` and lack of flexibility appear to be the primary cause of its poor performance, highlighting the need for agents that can balance exploitation with exploration.

\section{Conclusion}

We introduces the proposed \ourmethod{}agent for long-form video understanding, utilizing multi-granular search tools on constructed database for iterative search and reasoning over extensive video content. Our approach outperforms prior methods by adaptively integrating global browsing, clip search, and frame inspection, as demonstrated by state-of-the-art results on multiple benchmarks. Ablation studies confirm the effectiveness of our tool design, while analyses of reasoning model behavior provide insight into model reasoning patterns. Overall, our framework offers a scalable and flexible solution for comprehensive analysis of long videos.

\textbf{Limitations}. 
\label{sec:limitation}
While our agent significantly improves long video understanding, the iterative reasoning introduces higher computational overhead. In future work, we will explore more effective database construction and searching to reduce reasoning difficulty and thereby lower computational costs.

\newpage

\section*{Change Log}
\begin{itemize}[leftmargin=*,itemsep=2pt, parsep=0pt]
  \item \textbf{v1} (2025-05-23): Initial submission.
  \item \textbf{v2} (2025-05-28): Fixed the evaluation code to correctly account for answers enclosed in parentheses, resulting in consistently improved reported accuracy.
  \item \textbf{v3} (2025-07-24): GitHub repository link update.
\item \textbf{v4} (2025-11-03): Camera-ready version. Update Tab. 6 and 7 to discuss efficiency of our DVD.

\end{itemize}

\bibliographystyle{abbrvnat}
\bibliography{main}

\newpage

\appendix

 
\section{Evaluation Details}

\subsection{Prompts}

We describe the prompts used by different models within the proposed DVD, including: (1) Captioning and subject extraction model $M_{\text{database}}$ used in Multi-granular Video Database Construction (Tab.~\ref{tab:prompts_database}); (2) Reasoning model $M_{\text{reasoning}}$ used in Agentic Search and Answer (Tab.~\ref{tab:prompts_reasoning}); (3) Fine-grained detail querying model $M_{\text{tool}}$ used in the Frame Inspect tool (Tab.~\ref{tab:prompts_tool}); (4) event-centric global information in the Global Browse tool(Tab.~\ref{tab:prompts_global}). The JSON schema for tool use of OpenAI model are also provided in Tab.~\ref{tab:func_schema_part1} and Tab.~\ref{tab:func_schema_part2}.

\subsection{Cost}

We measure the average API cost of our agent when calling OpenAI o3 on LVBench. On average, DVD incurs a cost of \$0.213 with 0.15 million tokens per question with OpenAI API price at 2025-11-03.

\subsection{Azure OpenAI Service}

We utilize the LLM API through Azure OpenAI Service to evaluate both our proposed DVD agent and the OpenAI o3 model~\cite{openai2025o3}. However, we note two key limitations in the service that may affect performance: (1) a restriction on the maximum number of input images, and (2) a safety content filtering mechanism. This section discusses their impact and our mitigation strategies.

\subsubsection{Maximum Image Count Limitation}

Azure OpenAI Service imposes a limit of 50 images per request, which is significantly lower than the token limit allowed by Large VLMs we used. This restriction poses notable challenges in tasks involving long-form video understanding, as the model's comprehension capability strongly depends on the number of input frames provided. To overcome this limitation and deliver richer visual context within the imposed constraints, we adopt a strategy of combining multiple frames into a single composite image:

\textbf{For DVD}, we horizontally splice up to 5 frames into a single image. This allows us to input up to 250 frames (50 images × 5 frames). Note that we only use the 250 frames setting in event-centric global information gathering in the Global Browse tool while maximum 50 frames are used in the Frame Inspect tool.

\textbf{For OpenAI o3}, we use a 2×4 format—splicing up to 4 frames horizontally and stacking 2 such groups vertically per image. This enables our test condition of 256 frames by using 32 spliced images. We find that this configuration yields better performance than the 1×5 format used in DVD.

\subsubsection{API Content Filtering}
\label{sec:api}
Azure OpenAI's safety filtering mechanism occasionally misclassifies benign inputs as offensive, blocking some requests. This affects the performance of both OpenAI o3 and our DVD agent. For such filtered questions, we adopt different mitigation strategies to reduce its influence.

\textbf{For OpenAI o3}, when a query is blocked, we randomly select an answer from the remaining candidates. Across benchmarks, approximately 10\% to 25\% of queries are filtered, as shown in Tab.~\ref{tab:api_filtering}. To better assess true performance, we also report scores on the subset of unfiltered responses, where performance improves significantly.

\textbf{For DVD}, when VLM returns none due to the content filtering, we do not implement additional error handling for the agent. Instead, we let the agent make its own decisions to handle error.
It may invoke alternative tools or attempt to answer based on available information. If a captioning query is blocked, we leave the corresponding entry blank in the database. While this affects database completeness, the agent remains operational due to its adaptive behavior. Tab.~\ref{tab:api_filtering}.


\begin{table}[t]
    \caption{Statistics for content filtering in Azure OpenAI API.}
    \centering
    \renewcommand\arraystretch{1.15}
    \resizebox{0.94\linewidth}{!}{
    \begin{threeparttable}
        \begin{tabular}{l ccccc}
            \toprule
            \multirow{2}{*}{Methods}   & LVBench & \multicolumn{2}{c}{LongVideoBench (Val)}   & Video MME    & EgoSchema \\
            ~       & Overall & Overall & Long  & Long (w/o sub)  & Val \\
            \midrule
            \textbf{OpenAI o3}  & 57.1  & 66.7 & 59.6 & 64.7 & 63.2 \\
            Unfiltered Ratio & 83\%  & 90\% & 85\% & 83\% & 75\% \\
            Unfiltered Subset Score & 63.3 & 71.5 & 72.6 & 73.2 & 77.5 \\
            \midrule
            \textbf{Deep Video Discovery (Ours)} & 71.9 &  70.5 & 68.4  & 66.8 & 76.6  \\
            Unfiltered Ratio & 24\% & 61\% & 40\%  & 24\% & 54\%  \\
            Unfiltered Subset Score & 70.5 &  71.6 & 70.5 & 69.7 & 77.0  \\
            \bottomrule
        \end{tabular}
    \end{threeparttable}
    }
    \label{tab:api_filtering}
    \vspace{-2mm}
\end{table}

\section{More Results}

\subsection{Statistical Significance}

To assess the statistical significance of DVD, we evaluate it on LVBench (using auxiliary transcriptions) three times. It achieves an average score of 74.0 with a variance of 0.125, demonstrating the consistency and robustness of the proposed method.

\subsection{Case Study}

\label{sec:casestudy}

We provide illustrative examples of the tool-calling process for different agent behaviors described in Section 4.5 of the paper. All examples are from LVBench~\cite{wang2024lvbench}. Each example highlights how the agent selects actions and invokes tools to gather information for answering the question, and how each tool call retrieves relevant data from the database. In the tool responses, we highlight \textcolor{orange}{useful and correct information} in  \textcolor{orange}{orange} and \textcolor{purple}{useless or incorrect information}  in \textcolor{purple}{purple}.

\textbf{Global Browse Only} is shown in Fig.~\ref{fig:case-gob}. When asked about the type of video, the agent directly invokes Global Browse to retrieve global context. The subject-level browsing reveals many technical subjects, while the event-level browsing summarizes the key themes and categorizes the video as a technology documentary. Based on this information, the agent correctly classifies the video into a technology documentary.

\textbf{Simple Action} is shown in Fig.~\ref{fig:case-sa}. In this case, Global Browse incorrectly indicates that the mirror fragments are thrown away. To verify, the agent uses Clip Search to locate the time range of the mirror breaking event and then calls Frame Inspect for finer-grained analysis. It discovers that the fragments actually fly into the sky and thus selects the correct answer.

\textbf{Iterative Search} is shown in Fig.~\ref{fig:case-is}. This question involves multiple subjects (boy, dog, cat, bird, and man), making it more complex. Global Browse initially suggests that the man is chasing the dog. However, the first round of Clip Search and Frame Inspect indicates that he is chasing a bird. To resolve the ambiguity, the agent issues two more Clip Search queries for events involving "man chasing dog" and "man chasing bird". The search yields relevant and useful results only for the later, suggesting that the man is actually chasing the bird. To confirm the subject's identity, the agent checks for the presence of a mustache. Convinced that the mustached man is chasing the bird, the agent selects the correct option. Notably, when a Frame Inspect call fails to find relevant results in step 6, the agent can automatically ask another question to obtain an useful response.

\textbf{Frame Inspect Trap} is shown in Fig.~\ref{fig:case-fit}. This question targets an event at a specific timestamp. Around this time, the agent asks three consecutive questions. While Frame Inspect sometimes captures accurate information, it occasionally fails to interpret frames correctly, providing misleading responses. As a result, the agent cannot discern which information is reliable and ultimately answers incorrectly.

\textbf{Clip Search Trap} is shown in Fig.~\ref{fig:case-cst}. This example relies on identifying the correct time range using key terms like “protagonist” and “utility room.” However, the database contains few directly relevant captions or subjects. Despite the agent’s attempts to rephrase and refine its Clip Search queries, the tool fails to return relevant results, leading to an incorrect answer.

\section{Broader Impacts}

\subsection{Positive Impacts}

DVD facilitates natural language interaction with long-form video content, improving accessibility for users with visual or cognitive impairments. It can also enhance efficiency in media summarization and analysis, significantly reducing the need for manual review.

\subsection{Negative Impacts}

As DVD leverages large models (e.g., LLMs and VLMs) for captioning and reasoning, it inherits common concerns such as potential bias from training data, which may result in inaccurate or unfair interpretations of video content. Additionally, the agentic search process demands substantial computational resources, raising sustainability and accessibility concerns.

\begin{table*}[t]
\centering
\begin{minipage}{\textwidth}
\begin{lstlisting}
// system prompt
You are a helpful assistant.

// user prompt
There are consecutive frames from a video. BENCHMARK_CUSTOM_PLACEHOLDER Please understand the video clip with given corresponding transcript and historical subject information then output following JSON format with filled content:

Transcript of the current video clip:

[TRANSCRIPTS]

Historical subject information:

[SUBJECTS]

Output template:
{
  "clip_start_time": CLIP_START_TIME,
  "clip_end_time": CLIP_END_TIME,
  "new_subject_registry": { # add new subjects if any not in the historical subject information else leave it empty
    <new_subject_id>: {
      "name": <new_subject_name if the name is unknown please use 'unknown'>,
      "appearance": <a list of subject appearance description phase, including hair, face, clothes>,
      "identity": <a list of subject identity description shown in current video clip>,
      "first_seen": <timestamp>
    },
    ...
  },
  "clip_description": {
    "subjects_present": [<subject_id_list>],
    "captions": <a smooth and very detailed narration about the whole video clip>
  }
}
\end{lstlisting}
\end{minipage}
\caption{Prompts for captioning and subject extraction model $M_{\text{database}}$ in video database construction.}
\label{tab:prompts_database}
\end{table*}

\begin{table*}[h]
\centering
\begin{minipage}{\textwidth}
\begin{lstlisting}
// system prompt
You are a helpful assistant who can answer multistep questions by sequentially calling functions. Follow a pattern of THOUGHT (reason step-by-step about which function to call next), ACTION (call a function to as a next step towards the final answer), OBSERVATION (output of the function). Reason step by step which actions to take to get to the answer. 

Only call functions with arguments coming verbatim from the user or the output of other functions. Please keep going until the user's query is completely resolved, before ending your turn and yielding back to the user. Only terminate your turn when you are sure that the problem is solved. 

If you are not sure about file content or codebase structure pertaining to the user's request, use your tools to read files and gather the relevant information: do NOT guess or make up an answer. 

You MUST plan extensively before each function call, and reflect extensively on the outcomes of the previous function calls. DO NOT do this entire process by making function calls only, as this can impair your ability to solve the problem and think insightfully. 

The timestamp can be formatted as 'HH:MM:SS' or 'MM:SS'.

// user prompt
Carefully read timestamp and narration in the following script. Pay attention to the cause and sequence of events, the detail and movement of objects and the action and pose of persons.

You will be given a subject registry (which could be inaccurate) in this video and a video-level coarse understanding of the question. Please output your think process and use corresponding functions if you can not directly find answer from the given information.

+ CLIP_SEARCH_WITH_TIME_PLACEHOLDER If you need to find details without knowing the specific timespot, please use `ClipSearch`.

+ If the observed / searched information does not include accurate question-relevant content, e.g. 'unkown' name or other cases, use `FrameInspect` to ask a fine-grained question within a list of the specified time ranges (list[tuple[hhmmss, hhmmss]]).

+ If you can not make sure the answer by searched content, you should use `FrameInspect` to query related content in specified time ranges.

+ If you have found answer in a specific script, you should use `FrameInspect` to ask a fine-grained question to **confirm** it.

You can use multiple times of these functions to prgressively find the answer.

Based on your observations and returned result from functions, select the best option that accurately addresses the question.

Question: [question]

Answer with the option's letter from the given choices directly and only give the best option.

// history messages in the reasoning chain
[history messages]
\end{lstlisting}
\end{minipage}
\caption{Prompts for reasoning model $M_{\text{reasoning}}$.}
\label{tab:prompts_reasoning}
\end{table*}

\begin{table*}[h]
\centering
\begin{minipage}{\textwidth}
\begin{lstlisting}
// system prompt
You are a helpful assistant to answer questions.

// user prompt
Carefully watch the video frames. Pay attention to the cause and sequence of events, the detail and movement of objects and the action and pose of persons.

Based on your observations, if you find content that can answer the question, select the best option that accurately answers the question. If no relevant content is found within the given time range, return: `Error: Cannot find corresponding result in the given time range.`.

Question: [question]
\end{lstlisting}
\end{minipage}
\caption{Prompts for fine-grained detail query $M_{\text{tool}}$ in Frame Inspect tool.}
\label{tab:prompts_tool}
\end{table*}

\begin{table*}[h]
\centering
\begin{minipage}{\textwidth}
\begin{lstlisting}
// system prompt
You are a helpful assistant to answer questions.

// user prompt
Carefully watch the video frames. Pay attention to the cause and sequence of events, the detail and movement of objects and the action and pose of persons.

Based on your observations, provide detailed insights to help interpret the question and video. If you find content that answers the question, suggest a possible answer. If no relevant content is found in the sampled frames, returns `Error: Cannot find corresponding result in sampled frames.`. Format the answer into a single paragraph. 

Question: [question]
\end{lstlisting}
\end{minipage}
\caption{Prompts for query in Global Browse tool.}
\label{tab:prompts_global}
\end{table*}

\begin{table*}[h]
\centering
\begin{minipage}{\textwidth}
\begin{lstlisting}
[
    {
        "function": {
            "name": "global_browse",
            "description": "Obtain a subject registry and relevant information about the query from the whole video.\n\nReturns:\n    str: A formatted subject registry includes multiple identical attributes of subjects and a narration about relevant information about the query.\n",
            "parameters": {
                "properties": {
                    "database": {
                        "$ref": "#/$defs/NanoVectorDB",
                        "description": "The database object that supports querying with embeddings."
                    },
                    "query": {
                        "description": "A textual description of the question to search for.",
                        "type": "string"
                    }
                },
                "required": [
                    "database",
                    "query"
                ],
                "type": "object"
            }
        },
        "type": "function"
    },
    {
        "function": {
            "name": "clip_search",
            "description": "Searches for events in a video clip database based on a given event description and retrieves the top-k most relevant video clip captions.\n\nReturns:\n    str: A formatted string containing the concatenated captions of the searched video clip scripts.\n\nNotes:\n    - This function utilizes the OpenAI Embedding Service to generate embeddings for the input text.\n    - Use default values for `top_k` to limit the number of results returned.",
            "parameters": {
                "properties": {
                    "database": {
                        "$ref": "#/$defs/NanoVectorDB",
                        "description": "The database object that supports querying with embeddings."
                    },
                    "event_description": {
                        "description": "A textual description of the event to search for.",
                        "type": "string"
                    },
                    "top_k": {
                        "default": 16,
                        "description": "The maximum number of top results to retrieve. Just use the default value.",
                        "type": "integer"
                    }
                },
                "required": [
                    "database",
                    "event_description"
                ],
                "type": "object"
            }
        },
        "type": "function"
    }
]
\end{lstlisting}
\end{minipage}
\caption{Function JSON schema for OpenAI model's tool use. Part 1/2.}
\label{tab:func_schema_part1}
\end{table*}

\begin{table*}[h]
\centering
\begin{minipage}{\textwidth}
\begin{lstlisting}
[
    {
        "function": {
            "name": "frame_inspect",
            "description": "Crop the video frames based on the time ranges and ask the model a detailed question about the cropped video clips.\nReturns:\n    str: The model's response to the question. If no relevant content is found within the time range,\n         returns an error message: \"Error: Cannot find corresponding result in the given time range.\"",
            "parameters": {
                "properties": {
                    "database": {
                        "$ref": "#/$defs/NanoVectorDB",
                        "description": "The database containing video metadata. Must be an instance of NanoVectorDB."
                    },
                    "question": {
                        "description": "The specific detailed question to ask about the video content during the specified time ranges. No need to add time ranges in the question.",
                        "type": "string"
                    },
                    "time_ranges_hhmmss": {
                        "description": "A list of tuples containing start and end times in HH:MM:SS format. If the time range is longer than 50 seconds, the function samples 50 evenly distributed frames.  Otherwise, it uses all frames within the specified range.",
                        "items": {
                            "items": {},
                            "type": "array"
                        },
                        "type": "array"
                    }
                },
                "required": [
                    "database",
                    "question",
                    "time_ranges_hhmmss"
                ],
                "type": "object"
            }
        },
        "type": "function"
    },
    {
        "function": {
            "name": "answer",
            "description": "Call this function after confirming the answer of the user's question, and finish the conversation.",
            "parameters": {
                "properties": {
                    "answer": {
                        "description": "Answer to the user's question.",
                        "type": "string"
                    }
                },
                "required": [
                    "answer"
                ],
                "type": "object"
            }
        },
        "type": "function"
    }
]
\end{lstlisting}
\end{minipage}
\caption{Function JSON schema for OpenAI model's tool use. Part 2/2.}
\label{tab:func_schema_part2}
\end{table*}

\include{case_study/global_browse_only}
\include{case_study/simple_action}
\include{case_study/iterative_search}
\include{case_study/frame_inspect_trap}
\include{case_study/clip_search_trap}

\end{document}

%% file: case_study/global_browse_only.tex
\clearpage
\noindent
\begin{minipage}[t]{0.9\textwidth}
\centering
\resizebox{0.85\textwidth}{!}{
\begin{tcolorbox}[colback=white, colframe=gray!50!black, 
title=\textbf{Question}, boxrule=0.3mm, width=\textwidth, arc=2mm, auto outer arc=true]
Question: What type of video is this? \\
(A) A technology documentary\\
(B) A news report\\
(C) A travel documentary\\
(D) A travel vlog
\end{tcolorbox}
}
\end{minipage}

\begin{minipage}[t]{0.9\textwidth}
\centering
\resizebox{0.85\textwidth}{!}{
\begin{tcolorbox}[colback=orange!5!white, colframe=orange!75!black, 
title=\textbf{Agent (Step 1)}, boxrule=0.3mm, width=\textwidth, arc=2mm, auto outer arc=true]
Calling tool \textbf{Global Browse}.
\end{tcolorbox}
}
\end{minipage}

\hfill
\begin{minipage}[t]{0.9\textwidth}
\centering
\begin{adjustbox}{max width=0.85\textwidth}
\begin{tcolorbox}[
  colback=white, colframe=blue!30!gray,
  title=\textbf{Global Browse},  boxrule=0.3mm, width=\textwidth, arc=3mm, auto outer arc=true]
\textbf{Subjects Information : }
\begin{lstlisting}[basicstyle=\small, frame=none, backgroundcolor=\color{white}, breaklines=true, showstringspaces=false, escapeinside={(*@}{@*)}]
...
"subject_2": {
    "name": "industrial_robot",
    "appearance": ...
    "identity": [
        "(*@\textcolor{orange}{\mbox{Industrial automation robot}}@*), performing repetitive manufacturing or assembly line tasks"
    ]
},
...
"subject_10": {
    "name": "digital_face_scan_display",
    "appearance": ..., 
    "identity": [
        "Computer interface for (*@\textcolor{orange}{\mbox{facial recognition and identification}}@*), used for matching and analyzing faces in security or surveillance contexts"
    ]
},
...
"subject_21": {
    "name": "Pedro Domingos",
    "appearance": ..., 
    "identity": [
        "(*@\textcolor{orange}{\mbox{Academic expert and commentator}}@*), providing context on the Go match and public expectations"
    ]
},
...
\end{lstlisting}
\textbf{Event Information : }
\begin{lstlisting}[basicstyle=\small, frame=none, backgroundcolor=\color{white}, breaklines=true, showstringspaces=false, escapeinside={(*@}{@*)}]
The montage shows extended, polished coverage of artificial-intelligence themes: the AlphaGo challenge match and Google DeepMind branding, autonomous trucks on desert highways, ... . Therefore, the video is best classified as a (*@\textcolor{orange}{\mbox{technology documentary}}@*).
\end{lstlisting}
\end{tcolorbox}
\end{adjustbox}
\end{minipage}

\begin{minipage}[t]{0.9\textwidth}
\centering
\resizebox{0.85\textwidth}{!}{
\begin{tcolorbox}[colback=orange!5!white, colframe=orange!75!black, 
title=\textbf{Agent (Step 2)}, boxrule=0.3mm, width=\textwidth, arc=2mm, auto outer arc=true]
Executing \textbf{Asnwer}: (A) A technology documentary. $\to$ \textcolor{red}{Right}
\end{tcolorbox}
}
\vspace{3mm}
\captionof{figure}{Case study of \textbf{Global Browse Only} behavior.}
\label{fig:case-gob}
\end{minipage}

%% file: case_study/simple_action.tex
\clearpage
\noindent
\begin{minipage}[t]{0.9\textwidth}
\centering
\resizebox{0.85\textwidth}{!}{
\begin{tcolorbox}[colback=white, colframe=gray!50!black, 
title=\textbf{Question}, boxrule=0.3mm, width=\textwidth, arc=2mm, auto outer arc=true]
Question: What happens when the villain's mirror breaks?\\
(A) The mirror fragments automatically fuse on the ground into a whole mirror again\\
(B) The mirror fragments cut the villain's hand\\
(C) The mirror fragments are thrown away by the villains\\
(D) The mirror fragments fly into the sky
\end{tcolorbox}
}
\end{minipage}

\begin{minipage}[t]{0.9\textwidth}
\centering
\resizebox{0.85\textwidth}{!}{
\begin{tcolorbox}[colback=orange!5!white, colframe=orange!75!black, 
title=\textbf{Agent (Step 1)}, boxrule=0.3mm, width=\textwidth, arc=2mm, auto outer arc=true]
Calling tool \textbf{Global Browse}.
\end{tcolorbox}
}
\end{minipage}

\hfill
\begin{minipage}[t]{0.9\textwidth}
\centering
\begin{adjustbox}{max width=0.85\textwidth}
\begin{tcolorbox}[
  colback=white, colframe=blue!30!gray,
  title=\textbf{Global Browse},  boxrule=0.3mm, width=\textwidth, arc=3mm, auto outer arc=true]
\textbf{Subjects Information : }
\begin{lstlisting}[basicstyle=\small, frame=none, backgroundcolor=\color{white}, breaklines=true, showstringspaces=false, escapeinside={(*@}{@*)}]
...
"subject_18": {
    "name": "(*@\textcolor{orange}{\mbox{Baron Badblood}}@*)",
    "appearance": ..., 
    "identity": [
        "(*@\textcolor{orange}{\mbox{Villainous character}}@*), shown invoking his magic (*@\textcolor{orange}{\mbox{mirror}}@*) to reveal ..."
    ]
},
...
\end{lstlisting}
\textbf{Event Information : }
\begin{lstlisting}[basicstyle=\small, frame=none, backgroundcolor=\color{white}, breaklines=true, showstringspaces=false, escapeinside={(*@}{@*)}]
The scene shows the villain ...; the next shots focus on the villain angrily flinging the pieces aside so that the multicolored shards clatter to the ground and lie scattered around his lair; ... ; therefore (*@\textcolor{purple}{the mirror fragments are thrown away by the villains}@*).
\end{lstlisting}
\end{tcolorbox}
\end{adjustbox}
\end{minipage}

\begin{minipage}[t]{0.9\textwidth}
\centering
\resizebox{0.85\textwidth}{!}{
\begin{tcolorbox}[colback=orange!5!white, colframe=orange!75!black, 
title=\textbf{Agent (Step 2)}, boxrule=0.3mm, width=\textwidth, arc=2mm, auto outer arc=true]
Calling tool \textbf{Clip Search} with query "villain mirror breaks shards thrown away."
\end{tcolorbox}
}
\end{minipage}

\hfill
\begin{minipage}[t]{0.9\textwidth}
\centering
\begin{adjustbox}{max width=0.85\textwidth}
\begin{tcolorbox}[
  colback=white, colframe=blue!30!gray,
  title=\textbf{Clip Search},  boxrule=0.3mm, width=\textwidth, arc=3mm, auto outer arc=true]
\textbf{Searched Video Clip Scripts : }
\begin{lstlisting}[basicstyle=\small, frame=none, backgroundcolor=\color{white}, breaklines=true, showstringspaces=false, escapeinside={(*@}{@*)}]
[00:03:7.0, 00:03:20.0] ... Baron Badblood stands before a large, ornate mirror ...
...
[00:09:46.0, 00:10:6.0] ... (*@\textcolor{orange}{\mbox{Baron strikes the mirror}}@*) with a sword, causing it to shatter. The shards (*@\textcolor{orange}{\mbox{explode outward, tumbling and spinning through the air}}@*) as they scatter ...
...
[00:23:52.0, 00:24:4.0] ... The mirror shimmers with swirling pink light,  ...
\end{lstlisting}
\end{tcolorbox}
\end{adjustbox}
\end{minipage}

\begin{minipage}[t]{0.9\textwidth}
\centering
\resizebox{0.85\textwidth}{!}{
\begin{tcolorbox}[colback=orange!5!white, colframe=orange!75!black, 
title=\textbf{Agent (Step 3)}, boxrule=0.3mm, width=\textwidth, arc=2mm, auto outer arc=true]
Calling tool \textbf{Frame Inspect} at ["00:09:46", "00:10:06"]: "What do the broken pieces of the mirror do after Baron Badblood smashes it?"
\end{tcolorbox}
}
\end{minipage}

\hfill
\begin{minipage}[t]{0.9\textwidth}
\centering
\begin{adjustbox}{max width=0.85\textwidth}
\begin{tcolorbox}[
  colback=white, colframe=blue!30!gray,
  title=\textbf{Frame Inspect},  boxrule=0.3mm, width=\textwidth, arc=3mm, auto outer arc=true]
\textbf{Fine-grained Response : }
\begin{lstlisting}[basicstyle=\small, frame=none, backgroundcolor=\color{white}, breaklines=true, showstringspaces=false, escapeinside={(*@}{@*)}]
The fragments come to life: they quiver, gather into a swirling flock and fly out of the room, (*@\textcolor{orange}{\mbox{sailing away into the sky}}@*).
\end{lstlisting}
\end{tcolorbox}
\end{adjustbox}
\end{minipage}

\begin{minipage}[t]{0.9\textwidth}
\centering
\resizebox{0.85\textwidth}{!}{
\begin{tcolorbox}[colback=orange!5!white, colframe=orange!75!black, 
title=\textbf{Agent (Step 4)}, boxrule=0.3mm, width=\textwidth, arc=2mm, auto outer arc=true]
Executing \textbf{Asnwer}: (D) The mirror fragments fly into the sky.  $\to$ \textcolor{red}{Right}
\end{tcolorbox}
}
\vspace{3mm}
\captionof{figure}{Case study of \textbf{Simple Action} behavior.}
\label{fig:case-sa}
\end{minipage}

%% file: case_study/iterative_search.tex
\clearpage
\noindent
\begin{minipage}[t]{0.9\textwidth}
\centering
\resizebox{0.85\textwidth}{!}{
\begin{tcolorbox}[colback=white, colframe=gray!50!black, 
title=\textbf{Question}, boxrule=0.3mm, width=\textwidth, arc=2mm, auto outer arc=true]
Question: What does the man with the mustache do when boy, cat, and dog carrying treasure chest, shovel and pickaxe back?\\
(A) He chases a cat\\
(B) He chases a cow\\
(C) He chases a dog\\
(D) He chases a bird
\end{tcolorbox}
}
\end{minipage}

\begin{minipage}[t]{0.9\textwidth}
\centering
\resizebox{0.85\textwidth}{!}{
\begin{tcolorbox}[colback=orange!5!white, colframe=orange!75!black, 
title=\textbf{Agent (Step 1)}, boxrule=0.3mm, width=\textwidth, arc=2mm, auto outer arc=true]
Calling tool \textbf{Global Browse}.
\end{tcolorbox}
}
\end{minipage}

\hfill
\begin{minipage}[t]{0.9\textwidth}
\centering
\begin{adjustbox}{max width=0.85\textwidth}
\begin{tcolorbox}[
  colback=white, colframe=blue!30!gray,
  title=\textbf{Global Browse},  boxrule=0.3mm, width=\textwidth, arc=3mm, auto outer arc=true]
\textbf{Subjects Information : }...
\begin{lstlisting}[basicstyle=\small, frame=none, backgroundcolor=\color{white}, breaklines=true, showstringspaces=false, escapeinside={(*@}{@*)}]
\end{lstlisting}
\textbf{Event Information : }
\begin{lstlisting}[basicstyle=\small, frame=none, backgroundcolor=\color{white}, breaklines=true, showstringspaces=false, escapeinside={(*@}{@*)}]
Among the night-time forest shots the boy, cat and dog are shown tramping back along the path laden. The black-moustached postman suddenly rushes out and runs after Sharik the dog, trying to (*@\textcolor{purple}{grab the dog}@*), while the cat and boy keep hold of the chest and tools.
\end{lstlisting}
\end{tcolorbox}
\end{adjustbox}
\end{minipage}

\begin{minipage}[t]{0.9\textwidth}
\centering
\resizebox{0.85\textwidth}{!}{
\begin{tcolorbox}[colback=orange!5!white, colframe=orange!75!black, 
title=\textbf{Agent (Step 2)}, boxrule=0.3mm, width=\textwidth, arc=2mm, auto outer arc=true]
Calling tool \textbf{Clip Search} with query "man with mustache chases dog while boy cat carrying treasure chest".
\end{tcolorbox}
}
\end{minipage}

\hfill
\begin{minipage}[t]{0.9\textwidth}
\centering
\begin{adjustbox}{max width=0.85\textwidth}
\begin{tcolorbox}[
  colback=white, colframe=blue!30!gray,
  title=\textbf{Clip Search},  boxrule=0.3mm, width=\textwidth, arc=3mm, auto outer arc=true]
\textbf{Searched Video Clip Scripts : }
\begin{lstlisting}[basicstyle=\small, frame=none, backgroundcolor=\color{white}, breaklines=true, showstringspaces=false, escapeinside={(*@}{@*)}]
...
[00:08:0.0, 00:08:20.0] ... the (*@\textcolor{orange}{boy}@*) pushes a cart carrying a large, padlocked chest, followed closely by the (*@\textcolor{orange}{\mbox{cat and dog}}@*), each carrying a pickaxe and shovel ... Overhead, a (*@\textcolor{orange}{\mbox{black crow}}@*) soars across the sky ... Meanwhile, the tall thin (*@\textcolor{orange}{\mbox{man rushes after the group}}@*), ...
...
\end{lstlisting}
\end{tcolorbox}
\end{adjustbox}
\end{minipage}

\begin{minipage}[t]{0.9\textwidth}
\centering
\resizebox{0.85\textwidth}{!}{
\begin{tcolorbox}[colback=orange!5!white, colframe=orange!75!black, 
title=\textbf{Agent (Step 3)}, boxrule=0.3mm, width=\textwidth, arc=2mm, auto outer arc=true]
Calling tool \textbf{Frame Inspect} at ["00:08:00", "00:08:24"]: "Describe exactly what the tall thin adult man does as the boy, cat, and dog carry the treasure chest, shovel, and pickaxe back along the path. Is he chasing anyone specifically?"
\end{tcolorbox}
}
\end{minipage}

\hfill
\begin{minipage}[t]{0.9\textwidth}
\centering
\begin{adjustbox}{max width=0.85\textwidth}
\begin{tcolorbox}[
  colback=white, colframe=blue!30!gray,
  title=\textbf{Frame Inspect},  boxrule=0.3mm, width=\textwidth, arc=3mm, auto outer arc=true]
\textbf{Fine-grained Response : }
\begin{lstlisting}[basicstyle=\small, frame=none, backgroundcolor=\color{white}, breaklines=true, showstringspaces=false, escapeinside={(*@}{@*)}]
As the boy, cat, and dog come back, the man appears on the same path.

1. He is sprinting, coat and scarf flying, eyes fixed on the sky.
2. His arms reach up as he runs, grabbing at the air.
3. He (*@\textcolor{orange}{\mbox{never even looks at the boy, the animals or the chest}}@*); instead he is clearly trying to (*@\textcolor{orange}{\mbox{reach the black crow}}@*).
4. The man leaps and claws at the air, following the crow's flight.

So he is not chasing the boy, the cat, the dog or the treasure chest at all; he is chasing only the crow that is carrying the white object.
\end{lstlisting}
\end{tcolorbox}
\end{adjustbox}
\end{minipage}

\begin{minipage}[t]{0.9\textwidth}
\centering
\resizebox{0.85\textwidth}{!}{
\begin{tcolorbox}[colback=orange!5!white, colframe=orange!75!black, 
title=\textbf{Agent (Step 4)}, boxrule=0.3mm, width=\textwidth, arc=2mm, auto outer arc=true]
Calling tool \textbf{Clip Search} with query "mustached man chases dog".
\end{tcolorbox}
}
\end{minipage}

\hfill
\begin{minipage}[t]{0.9\textwidth}
\centering
\begin{adjustbox}{max width=0.85\textwidth}
\begin{tcolorbox}[
  colback=white, colframe=blue!30!gray,
  title=\textbf{Clip Search},  boxrule=0.3mm, width=\textwidth, arc=3mm, auto outer arc=true]
\textbf{Searched Video Clip Scripts : }
\begin{lstlisting}[basicstyle=\small, frame=none, backgroundcolor=\color{white}, breaklines=true, showstringspaces=false, escapeinside={(*@}{@*)}]
...
[00:33:12.0, 00:33:32.0] ... The cat darts away down the sandy road, desperate to escape, while the (*@\textcolor{purple}{dog}@*) quickly recovers and (*@\textcolor{purple}{\mbox{resumes the chase}}@*), bounding after the cat ...
...
\end{lstlisting}
\end{tcolorbox}
\end{adjustbox}
\end{minipage}

\begin{minipage}[t]{0.9\textwidth}
\centering
\resizebox{0.85\textwidth}{!}{
\begin{tcolorbox}[colback=orange!5!white, colframe=orange!75!black, 
title=\textbf{Agent (Step 5)}, boxrule=0.3mm, width=\textwidth, arc=2mm, auto outer arc=true]
Calling tool \textbf{Clip Search} with query "man with mustache chases crow".
\end{tcolorbox}
}
\end{minipage}

\hfill
\begin{minipage}[t]{0.9\textwidth}
\centering
\begin{adjustbox}{max width=0.85\textwidth}
\begin{tcolorbox}[
  colback=white, colframe=blue!30!gray,
  title=\textbf{Clip Search},  boxrule=0.3mm, width=\textwidth, arc=3mm, auto outer arc=true]
\textbf{Searched Video Clip Scripts : }
\begin{lstlisting}[basicstyle=\small, frame=none, backgroundcolor=\color{white}, breaklines=true, showstringspaces=false, escapeinside={(*@}{@*)}]
...
[00:08:28.0, 00:08:48.0] ... The (*@\textcolor{orange}{man}@*) points accusatorily at the boy and the crow, prompting the boy to hand over a shiny white object which (*@\textcolor{orange}{the crow had stolen earlier}@*) ...
...
\end{lstlisting}
\end{tcolorbox}
\end{adjustbox}
\end{minipage}

\begin{minipage}[t]{0.9\textwidth}
\centering
\resizebox{0.85\textwidth}{!}{
\begin{tcolorbox}[colback=orange!5!white, colframe=orange!75!black, 
title=\textbf{Agent (Step 6)}, boxrule=0.3mm, width=\textwidth, arc=2mm, auto outer arc=true]
Calling tool \textbf{Frame Inspect} at ["00:08:06", "00:08:14"]: "Does the man who is running after the crow have a visible mustache? Describe his facial hair."
\end{tcolorbox}
}
\end{minipage}

\hfill
\begin{minipage}[t]{0.9\textwidth}
\centering
\begin{adjustbox}{max width=0.85\textwidth}
\begin{tcolorbox}[
  colback=white, colframe=blue!30!gray,
  title=\textbf{Frame Inspect},  boxrule=0.3mm, width=\textwidth, arc=3mm, auto outer arc=true]
Cannot find corresponding result in the given time range.
\end{tcolorbox}
\end{adjustbox}
\end{minipage}

\begin{minipage}[t]{0.9\textwidth}
\centering
\resizebox{0.85\textwidth}{!}{
\begin{tcolorbox}[colback=orange!5!white, colframe=orange!75!black, 
title=\textbf{Agent (Step 7)}, boxrule=0.3mm, width=\textwidth, arc=2mm, auto outer arc=true]
Calling tool \textbf{Frame Inspect} at ["00:08:06", "00:08:14"]: "Does the tall thin man chasing the crow have a mustache, beard, or is he clean-shaven?"
\end{tcolorbox}
}
\end{minipage}

\hfill
\begin{minipage}[t]{0.9\textwidth}
\centering
\begin{adjustbox}{max width=0.85\textwidth}
\begin{tcolorbox}[
  colback=white, colframe=blue!30!gray,
  title=\textbf{Frame Inspect},  boxrule=0.3mm, width=\textwidth, arc=3mm, auto outer arc=true]
\textbf{Fine-grained Response : }
\begin{lstlisting}[basicstyle=\small, frame=none, backgroundcolor=\color{white}, breaklines=true, showstringspaces=false, escapeinside={(*@}{@*)}]
He has a (*@\textcolor{orange}{mustache}@*).
\end{lstlisting}
\end{tcolorbox}
\end{adjustbox}
\end{minipage}

\begin{minipage}[t]{0.9\textwidth}
\centering
\resizebox{0.85\textwidth}{!}{
\begin{tcolorbox}[colback=orange!5!white, colframe=orange!75!black, 
title=\textbf{Agent (Step 8)}, boxrule=0.3mm, width=\textwidth, arc=2mm, auto outer arc=true]
Executing \textbf{Asnwer}: (D) He chases a bird. $\to$ \textcolor{red}{Right}
\end{tcolorbox}
}
\vspace{3mm}
\captionof{figure}{Case study of \textbf{Iterative Search} behavior.}
\label{fig:case-is}
\end{minipage}

%% file: case_study/frame_inspect_trap.tex
\clearpage
\noindent
\begin{minipage}[t]{0.9\textwidth}
\centering
\resizebox{0.85\textwidth}{!}{
\begin{tcolorbox}[colback=white, colframe=gray!50!black, 
title=\textbf{Question}, boxrule=0.3mm, width=\textwidth, arc=2mm, auto outer arc=true]
Question: Why does the tall man cut off the chains at 21:49?\\
(A) To open the gate\\
(B) To blow out the candles\\
(C) To trigger a trap\\
(D) To help the character
\end{tcolorbox}
}
\end{minipage}

\begin{minipage}[t]{0.9\textwidth}
\centering
\resizebox{0.85\textwidth}{!}{
\begin{tcolorbox}[colback=orange!5!white, colframe=orange!75!black, 
title=\textbf{Agent (Step 1)}, boxrule=0.3mm, width=\textwidth, arc=2mm, auto outer arc=true]
Calling tool \textbf{Global Browse}.
\end{tcolorbox}
}
\end{minipage}

\hfill
\begin{minipage}[t]{0.9\textwidth}
\centering
\begin{adjustbox}{max width=0.85\textwidth}
\begin{tcolorbox}[
  colback=white, colframe=blue!30!gray,
  title=\textbf{Global Browse},  boxrule=0.3mm, width=\textwidth, arc=3mm, auto outer arc=true]
\textbf{Subjects Information : }...
\begin{lstlisting}[basicstyle=\small, frame=none, backgroundcolor=\color{white}, breaklines=true, showstringspaces=false, escapeinside={(*@}{@*)}]
\end{lstlisting}
\textbf{Event Information : }
\begin{lstlisting}[basicstyle=\small, frame=none, backgroundcolor=\color{white}, breaklines=true, showstringspaces=false, escapeinside={(*@}{@*)}]
The tall man steps up to a hanging length of chain that holds the barred door of a cell shut, raises his sword and slices clean through the links; the chain clatters free and the cell door can now be pushed open, allowing the (*@\textcolor{purple}{wounded protagonist inside to be supported out}@*).
\end{lstlisting}
\end{tcolorbox}
\end{adjustbox}
\end{minipage}

\begin{minipage}[t]{0.9\textwidth}
\centering
\resizebox{0.85\textwidth}{!}{
\begin{tcolorbox}[colback=orange!5!white, colframe=orange!75!black, 
title=\textbf{Agent (Step 2)}, boxrule=0.3mm, width=\textwidth, arc=2mm, auto outer arc=true]
Calling tool \textbf{Frame Inspect} at ["00:21:40", "00:21:58"]: "Why does the tall man cut off the chains?"
\end{tcolorbox}
}
\end{minipage}

\hfill
\begin{minipage}[t]{0.9\textwidth}
\centering
\begin{adjustbox}{max width=0.85\textwidth}
\begin{tcolorbox}[
  colback=white, colframe=blue!30!gray,
  title=\textbf{Frame Inspect},  boxrule=0.3mm, width=\textwidth, arc=3mm, auto outer arc=true]
\textbf{Fine-grained Response : }
\begin{lstlisting}[basicstyle=\small, frame=none, backgroundcolor=\color{white}, breaklines=true, showstringspaces=false, escapeinside={(*@}{@*)}]
He severs the ceiling chains so the (*@\textcolor{purple}{\mbox{candle-chandelier comes crashing down}}@*), plunging the corridor into darkness, (*@\textcolor{purple}{giving his companion a fighting chance to survive the attack}@*).
\end{lstlisting}
\end{tcolorbox}
\end{adjustbox}
\end{minipage}

\begin{minipage}[t]{0.9\textwidth}
\centering
\resizebox{0.85\textwidth}{!}{
\begin{tcolorbox}[colback=orange!5!white, colframe=orange!75!black, 
title=\textbf{Agent (Step 3)}, boxrule=0.3mm, width=\textwidth, arc=2mm, auto outer arc=true]
Calling tool \textbf{Frame Inspect} at ["00:21:45", "00:21:55"]: "What happens immediately after the tall man slices through the chains hanging from the ceiling?"
\end{tcolorbox}
}
\end{minipage}

\hfill
\begin{minipage}[t]{0.9\textwidth}
\centering
\begin{adjustbox}{max width=0.85\textwidth}
\begin{tcolorbox}[
  colback=white, colframe=blue!30!gray,
  title=\textbf{Frame Inspect},  boxrule=0.3mm, width=\textwidth, arc=3mm, auto outer arc=true]
\textbf{Fine-grained Response : }
\begin{lstlisting}[basicstyle=\small, frame=none, backgroundcolor=\color{white}, breaklines=true, showstringspaces=false, escapeinside={(*@}{@*)}]
The chain he cuts (*@\textcolor{orange}{\mbox{holds the candle-chandelier}}@*); as soon as it is severed, the chandelier comes crashing down from the ceiling onto the floor.
\end{lstlisting}
\end{tcolorbox}
\end{adjustbox}
\end{minipage}

\begin{minipage}[t]{0.9\textwidth}
\centering
\resizebox{0.85\textwidth}{!}{
\begin{tcolorbox}[colback=orange!5!white, colframe=orange!75!black, 
title=\textbf{Agent (Step 4)}, boxrule=0.3mm, width=\textwidth, arc=2mm, auto outer arc=true]
Calling tool \textbf{Frame Inspect} at ["00:21:40", "00:22:10"]: "Does the tall man ever cut a chain to open a gate or door around 21:49?"
\end{tcolorbox}
}
\end{minipage}

\hfill
\begin{minipage}[t]{0.9\textwidth}
\centering
\begin{adjustbox}{max width=0.85\textwidth}
\begin{tcolorbox}[
  colback=white, colframe=blue!30!gray,
  title=\textbf{Frame Inspect},  boxrule=0.3mm, width=\textwidth, arc=3mm, auto outer arc=true]
\textbf{Fine-grained Response : }
\begin{lstlisting}[basicstyle=\small, frame=none, backgroundcolor=\color{white}, breaklines=true, showstringspaces=false, escapeinside={(*@}{@*)}]
No.
\end{lstlisting}
\end{tcolorbox}
\end{adjustbox}
\end{minipage}

\begin{minipage}[t]{0.9\textwidth}
\centering
\resizebox{0.85\textwidth}{!}{
\begin{tcolorbox}[colback=orange!5!white, colframe=orange!75!black, 
title=\textbf{Agent (Step 5)}, boxrule=0.3mm, width=\textwidth, arc=2mm, auto outer arc=true]
Executing \textbf{Asnwer}: (C) To trigger a trap. $\to$ \textcolor{red}{Wrong}
\end{tcolorbox}
}
\vspace{3mm}
\captionof{figure}{Case study of \textbf{Frame Inspect Trap} behavior.}
\label{fig:case-fit}
\end{minipage}

%% file: case_study/clip_search_trap.tex
\clearpage
\noindent
\begin{minipage}[t]{0.9\textwidth}
\centering
\resizebox{0.85\textwidth}{!}{
\begin{tcolorbox}[colback=white, colframe=gray!50!black, 
title=\textbf{Question}, boxrule=0.3mm, width=\textwidth, arc=2mm, auto outer arc=true]
Question: What does the protagonist see through the window after she is taken to the utility room?\\
(A) A group of monks sitting cross-legged in the snow\\
(B) A group of citizens chatting together\\
(C) A group of warriors practicing swords\\
(D) A group of samurais eating
\end{tcolorbox}
}
\end{minipage}

\begin{minipage}[t]{0.9\textwidth}
\centering
\resizebox{0.85\textwidth}{!}{
\begin{tcolorbox}[colback=orange!5!white, colframe=orange!75!black, 
title=\textbf{Agent (Step 1)}, boxrule=0.3mm, width=\textwidth, arc=2mm, auto outer arc=true]
Calling tool \textbf{Global Browse}.
\end{tcolorbox}
}
\end{minipage}

\hfill
\begin{minipage}[t]{0.9\textwidth}
\centering
\begin{adjustbox}{max width=0.85\textwidth}
\begin{tcolorbox}[
  colback=white, colframe=blue!30!gray,
  title=\textbf{Global Browse},  boxrule=0.3mm, width=\textwidth, arc=3mm, auto outer arc=true]
\textbf{Subjects Information : }...
\begin{lstlisting}[basicstyle=\small, frame=none, backgroundcolor=\color{white}, breaklines=true, showstringspaces=false, escapeinside={(*@}{@*)}]
\end{lstlisting}
\textbf{Event Information : }
\begin{lstlisting}[basicstyle=\small, frame=none, backgroundcolor=\color{white}, breaklines=true, showstringspaces=false, escapeinside={(*@}{@*)}]
Inside the cramped utility room the heroine peers out through the lattice window and her eyes settle on the snow-covered courtyard where (*@\textcolor{purple}{\mbox{several monks sit cross-legged}}@*) and bare-chested despite the freezing weather, engaged in silent meditation.
\end{lstlisting}
\end{tcolorbox}
\end{adjustbox}
\end{minipage}

\begin{minipage}[t]{0.9\textwidth}
\centering
\resizebox{0.85\textwidth}{!}{
\begin{tcolorbox}[colback=orange!5!white, colframe=orange!75!black, 
title=\textbf{Agent (Step 2)}, boxrule=0.3mm, width=\textwidth, arc=2mm, auto outer arc=true]
Calling tool \textbf{Clip Search} with query "utility room window sees group monks cross-legged snow".
\end{tcolorbox}
}
\end{minipage}

\hfill
\begin{minipage}[t]{0.9\textwidth}
\centering
\begin{adjustbox}{max width=0.85\textwidth}
\begin{tcolorbox}[
  colback=white, colframe=blue!30!gray,
  title=\textbf{Clip Search},  boxrule=0.3mm, width=\textwidth, arc=3mm, auto outer arc=true]
\textbf{Searched Video Clip Scripts : }
\begin{lstlisting}[basicstyle=\small, frame=none, backgroundcolor=\color{white}, breaklines=true, showstringspaces=false, escapeinside={(*@}{@*)}]
...
[00:19:31.0, 00:19:42.0] ... As (*@\textcolor{purple}{soft snowflakes}@*) drift through the pale morning air, the solitary traveler appears on a frosted hill overlooking a vast, traditional Japanese town ...
\end{lstlisting}
\end{tcolorbox}
\end{adjustbox}
\end{minipage}

\begin{minipage}[t]{0.9\textwidth}
\centering
\resizebox{0.85\textwidth}{!}{
\begin{tcolorbox}[colback=orange!5!white, colframe=orange!75!black, 
title=\textbf{Agent (Step 3)}, boxrule=0.3mm, width=\textwidth, arc=2mm, auto outer arc=true]
Calling tool \textbf{Clip Search} with query "utility room protagonist looks through window sees".
\end{tcolorbox}
}
\end{minipage}

\hfill
\begin{minipage}[t]{0.9\textwidth}
\centering
\begin{adjustbox}{max width=0.85\textwidth}
\begin{tcolorbox}[
  colback=white, colframe=blue!30!gray,
  title=\textbf{Clip Search},  boxrule=0.3mm, width=\textwidth, arc=3mm, auto outer arc=true]
\textbf{Searched Video Clip Scripts : }
\begin{lstlisting}[basicstyle=\small, frame=none, backgroundcolor=\color{white}, breaklines=true, showstringspaces=false, escapeinside={(*@}{@*)}]
...
[00:02:2.0, 00:02:7.0] ... Dust motes drift lazily through beams of sunlight streaming from the paper (*@\textcolor{purple}{window}@*), highlighting the  wooden shelves lined with bowls ...
... 
\end{lstlisting}
\end{tcolorbox}
\end{adjustbox}
\end{minipage}

\begin{minipage}[t]{0.9\textwidth}
\centering
\resizebox{0.85\textwidth}{!}{
\begin{tcolorbox}[colback=orange!5!white, colframe=orange!75!black, 
title=\textbf{Agent (Step 4 - Step 14)}, boxrule=0.3mm, width=\textwidth, arc=2mm, auto outer arc=true]
Calling tool \textbf{Clip Search} with queries : \\
"utility room small storage taken escorted feed him utility", \\
"utility room window look out", \\
... \\
"storeroom small room snow see out window"
\end{tcolorbox}
}
\end{minipage}

\hfill
\begin{minipage}[t]{0.9\textwidth}
\centering
\begin{adjustbox}{max width=0.85\textwidth}
\begin{tcolorbox}[
  colback=white, colframe=blue!30!gray,
  title=\textbf{Clip Search},  boxrule=0.3mm, width=\textwidth, arc=3mm, auto outer arc=true]
\textbf{Searched Video Clip Scripts : }
\begin{lstlisting}[basicstyle=\small, frame=none, backgroundcolor=\color{white}, breaklines=true, showstringspaces=false, escapeinside={(*@}{@*)}]
...
\end{lstlisting}
\end{tcolorbox}
\end{adjustbox}
\end{minipage}

\begin{minipage}[t]{0.9\textwidth}
\centering
\resizebox{0.85\textwidth}{!}{
\begin{tcolorbox}[colback=orange!5!white, colframe=orange!75!black, 
title=\textbf{Agent (Step 15)}, boxrule=0.3mm, width=\textwidth, arc=2mm, auto outer arc=true]
Calling tool \textbf{Clip Search} with query "utility room".
\end{tcolorbox}
}
\end{minipage}

\hfill
\begin{minipage}[t]{0.9\textwidth}
\centering
\begin{adjustbox}{max width=0.85\textwidth}
\begin{tcolorbox}[
  colback=white, colframe=blue!30!gray,
  title=\textbf{Clip Search},  boxrule=0.3mm, width=\textwidth, arc=3mm, auto outer arc=true]
\textbf{Searched Video Clip Scripts : }
\begin{lstlisting}[basicstyle=\small, frame=none, backgroundcolor=\color{white}, breaklines=true, showstringspaces=false, escapeinside={(*@}{@*)}]
...
[00:17:32.0, 00:17:49.0] ... The setting is dim and somber, likely a (*@\textcolor{purple}{\mbox{shelter or workshop}}@*) with stone flooring and faint light filtering in from the outside ...
... 
\end{lstlisting}
\end{tcolorbox}
\end{adjustbox}
\end{minipage}

\begin{minipage}[t]{0.9\textwidth}
\centering
\resizebox{0.85\textwidth}{!}{
\begin{tcolorbox}[colback=orange!5!white, colframe=orange!75!black, 
title=\textbf{Agent (Step 16)}, boxrule=0.3mm, width=\textwidth, arc=2mm, auto outer arc=true]
Executing \textbf{Asnwer}: (A) A group of monks sitting cross-legged in the snow. $\to$ \textcolor{red}{Wrong}
\end{tcolorbox}
}
\vspace{3mm}
\captionof{figure}{Case study of \textbf{Clip Search Trap} behavior.}
\label{fig:case-cst}
\end{minipage}